\documentclass[11pt]{article}

\usepackage[]{acl}

\usepackage{times}
\usepackage{latexsym}
\usepackage{subcaption}
\usepackage{makecell}
\usepackage{booktabs}
\usepackage{siunitx}
\usepackage[T1]{fontenc}

\usepackage[utf8]{inputenc}

\usepackage{microtype}

\usepackage{inconsolata}

\usepackage{graphicx}
\usepackage{xcolor}
\usepackage{xspace}
\usepackage{booktabs}
\usepackage{multirow}
\usepackage{tikz}
\usepackage{pgfplots}
\pgfplotsset{compat=1.18}
\usepackage{xcolor}
\usepackage{multicol}
\usepackage{caption}
\usepackage{enumitem}
\usepackage{gradient-text}

\usepackage{amsmath}
\definecolor{DeepBlue}{HTML}{005CFF}       
\definecolor{Blue20}{RGB}{152,202,255}     
\usepgfplotslibrary{groupplots}

\title{\ours: Branch-Point Data Generation for GUI Agents}

\definecolor{YaleBlue}{RGB}{16, 42, 86}
\definecolor{UNCBlue}{RGB}{75, 156, 211}

\newcommand{\Unc}{\hspace{.1em}^{\textcolor{UNCBlue}{\boldsymbol{C}}}}

\newcommand{\Yale}{\hspace{.1em}^{\textcolor{YaleBlue}{\boldsymbol{Y}}}}

\newcommand{\huggingface}{\raisebox{-1.5pt}{\includegraphics[height=1.05em]{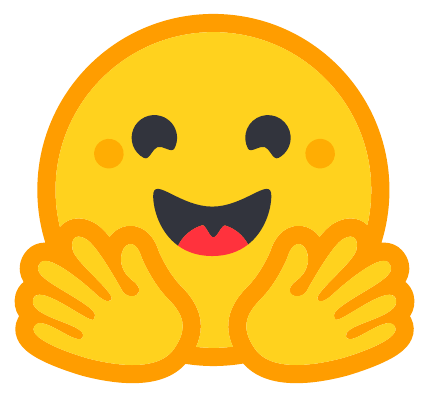}}\xspace}
\newcommand{\github}{\raisebox{-1.5pt}{\includegraphics[height=1.05em]{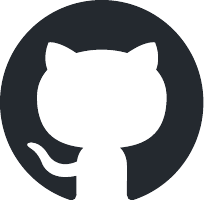}}\xspace}

\author{
Jinbiao Wei~$\Yale$ \quad Yilun Zhao\thanks{Correspondence: Yilun Zhao (\texttt{yilun.zhao@yale.edu})}$\Yale$ \quad 
Kangqi Ni$\Unc$ \quad Arman Cohan$\Yale$ \vspace{4pt}\\
$\Yale$~Yale University \quad $\Unc$~University of North Carolina at Chapel Hill
}

\newcommand{\ours}{%
  \textsc{\gradientRGB{Anchor}{63,93,203}{10,10,80}}%
  \xspace}

\begin{document}
\maketitle
\begin{abstract}
End-to-end GUI agents for real desktop environments require large amounts of high-quality interaction data, yet collecting human demonstrations is expensive and existing synthetic pipelines often suffer from limited task diversity or noisy, goal-drifting trajectories. We present a trajectory expansion framework \ours that bootstraps scalable desktop supervision from a small set of verified seed demonstrations. Starting from each seed, we identify branch points that correspond to meaningful state changes and propose new, state-grounded task variants conditioned on the current GUI context. An executing agent then follows the proposed instructions to generate new trajectories, while a verifier enforces task completion via state-aware checks and trajectory-level consistency. To improve supervision quality, we further apply task-conditioned step-level filtering to remove ungrounded actions and denoise post-branch segments to maintain coherent intent. Experiments on standard desktop benchmarks, OSWorld and WindowsAgentArena, show that models fine-tuned on our expanded corpus achieve consistent improvements over zero-shot agents and representative synthesis baselines, and generalize across applications and operating systems.

\begin{small}
\begin{center}
\begin{tabular}{cll}
\huggingface & \textbf{Data \& Model} & \href{https://huggingface.co/collections/yale-nlp/anchor} {\path{yale-nlp/Anchor}}\\
\github & \textbf{Code} & \href{https://github.com/yale-nlp/Anchor}{\path{yale-nlp/Anchor}}\\
\end{tabular}
\end{center} 
\end{small}
\vspace{5pt}
\end{abstract}

\section{Introduction}
The rapid advancements in large-scale Vision-Language Models (VLMs) and Large Language Models (LLMs) have opened up a new frontier for autonomous agents, particularly in the domain of interacting with Graphical User Interfaces (GUIs) ~\cite{su-etal-2024-language,qin2025ui,agasheagent}. These GUI agents aim to bridge visual perception, linguistic understanding, and physical action to autonomously complete complex, multi-step tasks across diverse interfaces, including the web ~\cite{dengmind2web,zhou2023webarena,zheng2024gpt}, desktop ~\cite{wu2024copilot} and mobile~\cite{rawlesandroidinthewild,yan2023gpt,gan2026android} platforms. This capability promises to unlock end-to-end automation for real-world processes. However, realizing this potential is fundamentally dependent on the availability of high-quality, large-scale training data.

Existing GUI benchmarks and agent datasets rely heavily on human-annotated trajectories~\cite{li2024effects, lu2024weblinx}, where experts carefully specify tasks, record step-by-step actions, and validate success. While such data is often high quality, it is extremely labor-intensive and difficult to scale across the combinatorial space of applications, configurations, and task goals. Generating datasets that map complex task specifications to successful, granular action trajectories across diverse GUI environments thus remains a significant bottleneck.
To mitigate the scarcity and cost of human demonstrations, recent work has explored scalable GUI-trajectory synthesis via several paradigms, including goal-conditioned task-driven generation, interaction-driven exploration with retroactive task inference, and tutorial-based replay (see \S\ref{sec:related_data_synthesis}). 
However, existing pipelines often struggle to produce long-horizon, high-signal trajectories and have largely concentrated on web/mobile settings, leaving desktop trajectory generation comparatively underexplored.

To bridge these gaps, we propose \ours, a trajectory expansion pipeline that systematically explores alternative branches around each seed trajectory. We start from a small set of tasks with strictly verified gold trajectories in desktop environments, and identify branch points along each trajectory where the UI reveals new affordances. 
At each branch point, we synthesize new tasks that diverge in goal while iteratively refining the task specification during execution.  Finally, we summarize each rollout into a concise task description and apply a verifier to confirm successful completion.

Experiments on two realistic desktop benchmarks, OSWorld~\cite{xie2024osworld} and WindowsAgentArena~\cite{bonatti2024windows}, demonstrate the effectiveness of \ours. Fine-tuning on trajectories generated by our branch-point expansion yields consistent gains across all evaluated backbones and all operating systems, outperforming representative synthesis baselines, and improving robustness across diverse applications and long-horizon workflows. These results suggest that scalable, state-grounded trajectory expansion is a practical route for turning general-purpose multimodal models into more capable desktop GUI agents. Our primary contributions are as follows:
\begin{itemize}[leftmargin=*]
    \item We develop a desktop GUI data generation and quality-control pipeline that produces \emph{high-diversity} and \emph{high-fidelity} trajectories while relying less on model perfection.
    \item We demonstrate the value of the resulting dataset by fine-tuning different models, which achieve strong performance on challenging desktop benchmarks and consistently outperform competitive baselines.
    \item We provide systematic ablations and diagnostic analyses to pinpoint what contributes most to the improvements and to better understand remaining failure modes.
\end{itemize}

\begin{figure*}[t]
  \centering
  \includegraphics[width=\textwidth]{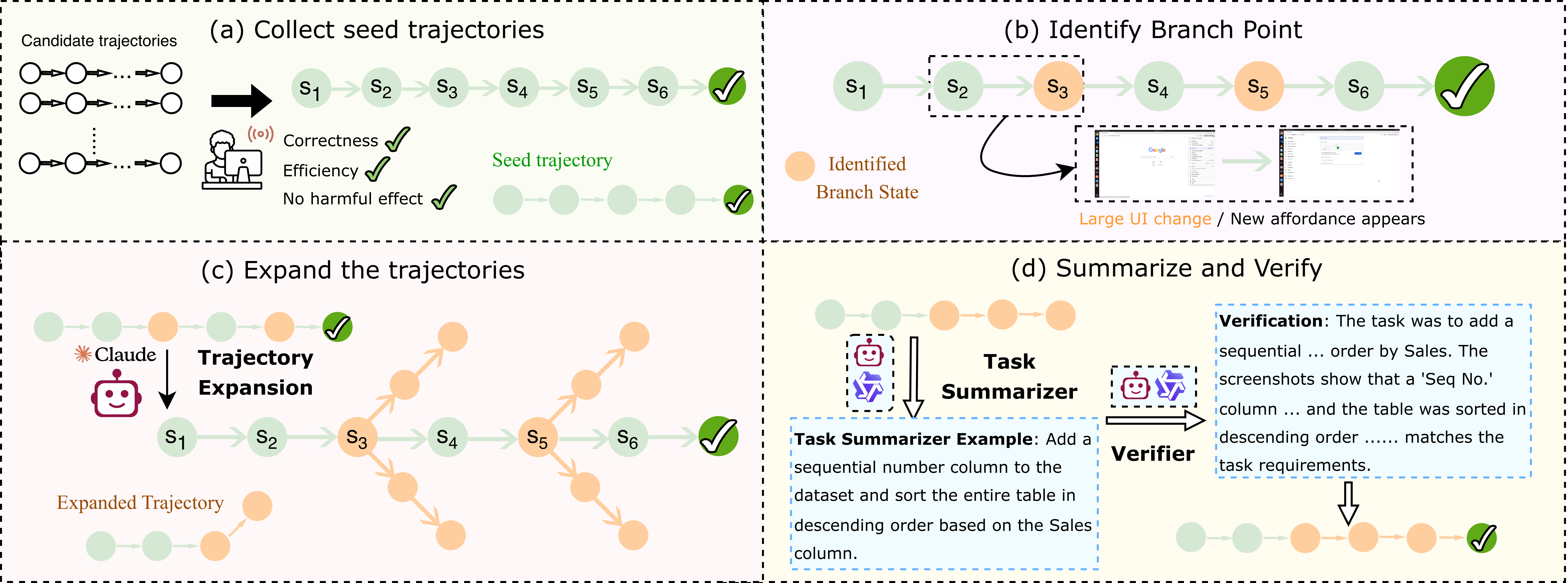}
  \caption{Trajectory generation pipeline. (a) Human annotators collect seed trajectories and retain only high-quality demonstrations. (b) We identify branch points along each seed where meaningful UI state changes reveal new affordances. (c) We then expand each seed by branching into diverse task instructions and executing the resulting trajectories. (d) Finally, a task summarizer produces a task description, and a verifier checks task completion.}
  \label{fig:pipeline}
\end{figure*}

\section{Related Work}
GUI agents iteratively observe screens and execute low-level actions. With foundation models closing the loop between perception and control, designs are converging on native computer-use frameworks (e.g., UI-TARS~\cite{qin2025ui}, Claude Computer Use~\cite{anthropic2024computeruse}, Agent S~\cite{agasheagent}). Scalable supervision remains a key bottleneck, motivating trajectory synthesis pipelines.
\label{sec:related_data_synthesis} 
\paragraph{Tutorial-Based Data Synthesis approaches} such as AgentTrek~\cite{xu2024agenttrek} and TongUI~\cite{zhang2025tongui} automatically crawl web tutorials or documentation, parse them into task specifications and step-by-step instructions, and then ask an agent to replay these instructions in an executable environment. Their coverage is fundamentally limited by the availability and diversity of tutorials. 

\paragraph{Task-Driven Data Synthesis approaches} focus on trajectory synthesis from explicit task specifications. For example, Explorer~\cite{pahuja2025explorer} uses an LLM to both formulate task intents and carry out their execution. A key limitation is that the resulting corpus is tightly coupled to the executor's competence: failures truncate rollouts and bias data toward easier tasks, while task diversity is bounded by what the proposal model knows about the UI and its affordances. In contrast, \ours reduces dependence on model competence by expanding around seed demonstrations and branches at UI-defined decision states, reducing reliance on proposing entirely new tasks from scratch and enabling systematic diversity from reliable prefixes. 
\paragraph{Interaction-Driven Data Synthesis approaches} push in the opposite direction, starting from free-form exploration without predefined tasks. OS-Genesis~\cite{sun2025genesis} and AutoPlay~\cite{ramrakhya2025scaling} for example, let an agent interact step-by-step with desktop and mobile environments such as AndroidWorld~\cite{rawles2024androidworld} and WebArena~\cite{zhou2023webarena}. GUI-ReWalk~\cite{lin2025guirewalkmassivedatageneration} extends this idea with a multi-stage framework: exploration to increase coverage, followed by latent-goal inference to assemble long-horizon workflows. These methods improve state coverage, but unconstrained exploration can yield either short, low-level traces or long segments of low-signal wandering, which are difficult to distill into meaningful long-horizon supervision. \ours avoids free-form wandering by initiating each rollout from a visually rich branch state with an explicit objective, producing longer yet goal-directed trajectories.

\section{GUI Agent Trajectory Expansion}
In this section, we detail \ours pipeline for synthesizing high-quality GUI agent trajectories by systematically expanding around a small set of high-quality seed demonstrations. 
\subsection{Pipeline Design Principles}
We design the trajectory expansion pipeline around the following three guiding principles.
(i) \textbf{Systematic diversity and coverage:} diversity comes from enumerating alternative, goal-directed branches at multiple UI-defined decision points, yielding broad state--action coverage without relying on unconstrained exploration; 
(ii) \textbf{Less wandering and higher signal:} because each rollout is initiated from a concrete branch point with an explicit objective, trajectories are less likely to degenerate into short, low-level clicks or long stretches of aimless navigation that are difficult to distill into meaningful multi-step tasks; (iii) \textbf{Reduced dependence on model perfection}: exploration is anchored by a small set of high-quality trajectories and the agent is only required to solve localized sub-tasks around them; we additionally apply step-level filtering and intention-consistency denoising to discard noisy or task-irrelevant steps. 
In the following subsection, we describe the concrete data generation procedure that instantiates these principles. \autoref{fig:pipeline} shows an overview of the trajectory generation pipeline.

\subsection{Data Generation Recipe}
\label{sec:data-generation}

We design a trajectory expansion pipeline that leverages existing GUI agent datasets with executable environments and a small set of high-quality seed trajectories. Given an
environment $\mathcal{E}$ and a task $T$, a gold trajectory is a sequence
$\tau = (s_0, a_0, s_1, \dots, s_{k})$ of GUI states $s_k$ and actions $a_{k-1}$ that successfully
completes $T$.

\paragraph{Seed trajectory collection.}
We collect seed trajectories directly from the interactive desktop environments of OSWorld~\cite{xie2024osworld} and WindowsAgentArena~\cite{bonatti2024windows}. Specifically, we sample 117 task environments from OSWorld and 51 task environments from WindowsAgentArena for data generation. Because our branching mechanism builds large families of trajectories around these seeds, it is crucial that they correspond to reliable and efficient solutions rather than noisy or suboptimal behavior. For each sampled task, we then collect a single high-quality \emph{seed trajectory} that will serve as the root for our branching procedure as follows. We first obtain one or more successful runs for the task using strong agents, and then perform a dedicated human validation pass over each candidate trajectory. Annotators (details in \autoref{app:annotators}) are instructed to (i) verify that the final GUI state fully satisfies the natural-language instruction, (ii) ensure that the sequence of actions is \emph{efficient}, in the sense that it does not contain obvious detours or redundant UI operations, and (iii) check that the trajectory does not introduce harmful side effects (e.g., deleting unrelated files or changing global settings). Trajectories that fail any of these checks are discarded. When multiple candidates pass, we retain the shortest one (i.e., the most efficient successful run), and treat this as one of the seed trajectories.

\paragraph{Branch-point identification.}
Given a seed trajectory $\tau$, we define a \emph{branch point} as a time step
$t$ such that we may stop the original task at state $s_t$ and initiate one or more new tasks
that all share the prefix $(s_0, a_0, \dots, s_t)$.  For each seed trajectory $\tau$, we use GPT-5.1
to identify promising \emph{branch points}. Intuitively, a step $t$ is a suitable branch point when:
(i) the user interface undergoes a substantial change (e.g., a new window, a panel
appears), or (ii) additional content becomes visible (e.g., after scrolling to reveal more of the page or pasting new text). This definition treats branch points as UI-defined decision nodes: the current screen exposes new affordances or intermediate artifacts, many distinct user-level goals can be pursued from the same verified prefix, enabling systematic diversity while avoiding unconstrained exploration. For each trajectory, we select $3$ -- $5$ such branch points, from which new tasks can naturally be defined.

\paragraph{Branch task proposal.}
For each branch point $(T, \tau, s_t)$, we first feed the trajectory prefix
$(s_0, a_0, \dots, s_t)$ to GPT-5.1 and ask it to produce a concise summary of the
user’s progress and the UI changes up to state $s_t$. We then prompt the model with this summary together with the current GUI
state $s_t$ to propose one or more new task descriptions $\{T'_{t,j}\}_j$. These
tasks are required to (i) be grounded in the visible interface at $s_t$ and
(ii) follow naturally from the preceding actions.

\paragraph{Task execution.}
For every proposed task $T'_{t,j}$, we replay the environment to state $s_t$ and invoke a
state-of-the-art GUI agent to complete the task, obtaining a candidate trajectory
$\tau'_{t,j}$. During execution, we let a model refine the task description on the fly whenever the agent’s actions drift from the original specification   (e.g., Clicking 'Font 19' option instead of the requested 'Font 20') or the environment state is incompatible with it (e.g., a referenced file is missing). In such cases, the model revises $T'_{t,j}$ into a feasible, semantically close variant that, in most instances, reinterprets even imperfect behavior as a coherent task.

\paragraph{Trajectory summarization.}
After the GUI agent terminates, we obtain a candidate trajectory $\hat{\tau}_{t,j} = (s_0, a_0, \dots, s_t, a_t, \dots, s_{K'})$ for task $T'_{t,j}$. We then apply a summarizer to produce a concise, high-level description $\hat{T}_{t,j}$ of the overall task, abstracting away low-level UI operations while capturing the user-level goal.

\paragraph{Trajectory verification.}
To ensure data quality, we employ a verifier that receives
$(\hat{T}_{t,j}, \hat{\tau}_{t,j})$ and judges whether the final GUI state satisfies the task
specification. We retain a trajectory only if (i) the GUI agent explicitly indicates task completion, and (ii) the verifier
classifies the trajectory as successful. Trajectories that are incoherent, incomplete, or misaligned with the final task description are discarded.

\subsection{Step-Level Filtering and Denoising}
\label{sec:step_denoising}
While trajectory-level verification removes failed rollouts, even successful synthetic trajectories can still contain step-level noise and intent drift that dilute the supervision signal. We therefore introduce two complementary step-level quality controls to retain only steps that are consistent with each task.
\paragraph{Task-conditioned reasoning filtering for shared prefix steps.}
A single prefix of low-level actions can be shared by many descendant tasks, but the intent
behind each action in that prefix can differ across tasks. For example, a prefix may open the file manager, navigate to \texttt{Downloads}, and click \texttt{report.pdf}; this identical sequence can be part of distinct descendant tasks such as \emph{attach the report to an email}, \emph{rename the report for archival}, or \emph{move the report into a project folder}. To capture these task-specific
interpretations---and to avoid supervising steps that are inconsistent with a particular
task---we enrich each trajectory with step-level reasoning that is conditioned on the
downstream task rather than on the action alone.
For each step $k \le t$ in the prefix shared by a branch point, and for each descendant task
$T'_{t,j}$ that inherits this prefix, we prompt an LLM with:
(i) the task description $T'_{t,j}$,
(ii) the interaction history up to state $s_k$, and
(iii) the current screenshot at step $k$.
The model is asked to sample $M$ candidate action--reason pairs
$\{(\tilde{a}_k^{(m)}, \tilde{r}_k^{(m)})\}_{m=1}^M$ that are plausible given this context.
For each candidate, we then query the LLM again with the pre- and post-action screenshots
$(s_k, s_{k+1})$ and the proposed action $\tilde{a}_k^{(m)}$, asking whether this action
is consistent with the observed visual change.
If at least one candidate is judged consistent, we select the rationale
$\tilde{r}_k^{(m^\ast)}$ whose action $\tilde{a}_k^{(m^\ast)}$ both (a) matches the executed
action $a_k$ in the gold trajectory and (b) is visually compatible with the transition
from $s_k$ to $s_{k+1}$.
The selected rationale is attached as the step-level explanation for that $(T'_{t,j}, k)$
pair.
If none of the $M$ proposals are consistent with the observed UI change, we treat this
$(T'_{t,j}, k)$ pair as task-irrelevant or noisy and simply drop it from supervision for
task $T'_{t,j}$.
In our experiments we set $M=10$.
This procedure yields diverse, task-conditioned reasoning for shared prefix steps while
filtering out steps that are not meaningfully grounded for a particular descendant task.

\paragraph{Post-branch intention-consistency denoising.}
Even after the branch point, synthetic trajectories steps can contain local noise such as misclicks, unnecessary waits, or transient detours that the agent later recovers from. We do not want the model to imitate these idiosyncratic behaviors. To this end, for each step $k > t$ we run an intention-consistency check. Given the task description $T'_{t,j}$, the interaction history up to $s_k$, the pre- and
post-action screenshots $(s_k, s_{k+1})$, and the logged action $a_k$, we prompt the LLM to (i) verify that the proposed action is reasonable in the current context and (ii) ensure that this action, in turn, matches the observed visual change between $s_k$ and $s_{k+1}$.
If this consistency test fails, we mark step $k$ as noisy and exclude it from training, while still retaining later steps in the same trajectory that pass the filter. Intuitively, this step-level denoising prevents the model from overfitting to accidental, low-signal UI operations, while preserving supervision on deliberate actions and the subsequent corrections that demonstrate how to recover from mistakes.

\begin{table}[!t]
\centering
\small
\begin{tabular}{@{}lc@{}}
\toprule
\bfseries Metric                        & \bfseries Value \\ \midrule
\# Total Ubuntu trajectories                         & \num{1174}  \\ 
\# Total Windows trajectories                    & \num{603}   \\ 
Average steps per trajectory          & \num{17.24}   \\  
\midrule
\# Tokens           & \num{2.3}M  \\
\# Images           & \num{30}K\\ \midrule
Cost per successful trajectory         & \$\num{0.47} \\
\bottomrule
\end{tabular}
\caption{Dataset statistics. The number of trajectories, average steps per trajectory, and number of tokens, and images correspond to the successful trajectories.
}
\label{tab:data_stat}
\end{table}

\subsection{Analysis of Synthetic Trajectories}\label{sec:data_analysis}

Our trajectory synthesis pipeline can be adopted with any high-capability models. In practice, we use Claude Sonnet 4.5~\cite{anthropic2025claude_sonnet_45} for trajectory execution, GPT-5.1~\cite{openai2025gpt51} for task proposition and replay verification, and Qwen3-VL-32B~\cite{bai2025qwen3vl} for step-level filtering, task summarization, and verification. 
Our final training corpus consists of 1,777 successful desktop trajectories spanning two OS platforms: 1,174 Ubuntu trajectories and 603 Windows trajectories at an average cost of \$0.47 per successful trajectory, as shown in \autoref{tab:data_stat}. In terms of interaction horizon, trajectories in our corpus average 17.24 steps, substantially longer than prior GUI synthesis pipelines (\autoref{tab:length}). Importantly, this increased horizon does not arise from unconstrained exploration or idle UI wandering, but reflects structured, goal-directed multi-stage workflows such as navigating hierarchical settings menus, configuring multi-field dialogs. Our branching-based expansion creates families of tasks that share valid GUI prefixes but diverge in deeper intent, yielding longer yet goal-directed trajectories aligned with realistic interface flows—unlike prior synthetic corpora dominated by short, single-panel interactions. Despite the increased horizon, step-level intention-consistency filtering ensures that the added depth does not introduce low-signal wandering actions, preserving the overall quality and learnability of the resulting supervision. All these data will be made public to accelerate future research.

\begin{table}[!t]
\centering
\small
\begin{tabular}{lc}
\toprule
\bfseries Model     & \bfseries Avg Trajectory Len\ \\ \midrule
Explorer \cite{pahuja2025explorer} & \num{7.7} \\
OS-Genesis \cite{sun2025genesis} & \num{5.6}     \\
\bfseries \ours & \num{17.24}    \\ 
\bottomrule
\end{tabular}
\caption{Trajectory length comparison with representative prior GUI synthesis pipelines; since previous data synthesis work focus on mobile/web GUIs, we do not directly evaluate them as baselines in our experiments; instead, we adopt their high-level ideas and implement a version on our own.}

\label{tab:length}
\end{table}

\begin{table*}[ht]
\centering
\resizebox{0.98\linewidth}{!}{
\begin{tabular}{llccccccccccc}
\toprule
\textbf{Model} & \textbf{Strategies} & \textbf{Chrome} & \textbf{GIMP} & \textbf{Calc} & \textbf{Impress} & \textbf{Multi-Apps} & \textbf{Writer} & \textbf{OS} & \textbf{Thunderbird} & \textbf{VLC} & \textbf{VS Code} & \textbf{Overall} \\
\midrule

\multirow{4}{*}{GLM4.1V-9B}
& Zero-Shot    & 3.33  & 0.00 & 0.00 & 0.00 & 0.00 & 0.00 & 0.00 & 0.00 & 0.00 & 0.00 & 0.47 \\
& Task-Driven  & 6.67  & 0.00 & \textbf{3.45} & 3.70 & 2.86 & 7.69 & \textbf{9.09} & \textbf{12.50} & 0.00 & \textbf{18.18} & 5.14 \\
& Human Data   & 6.67  & \textbf{12.50} & 0.00 & 3.70 & \textbf{4.29} & 7.69 & \textbf{9.09} & 0.00 & 0.00 & \textbf{18.18} & 5.14 \\
\cmidrule(lr){2-13}
& \ours   & \textbf{10.00} & \textbf{12.50} & \textbf{3.45} & \textbf{7.41} & 2.85 & \textbf{15.38} & 0.00 & \textbf{12.50} & \textbf{14.29} & 9.09 & \textbf{7.01} \\
\midrule

\multirow{4}{*}{Qwen2.5-VL-7B}
& Zero-Shot    & 3.33  & 0.00 & 0.00 & 0.00 & 1.43 & 0.00 & 0.00 & \textbf{0.00} & 0.00 & 0.00 & 0.93 \\
& Task-Driven  & \textbf{10.00} & 12.50 & 0.00 & 3.70 & 2.86 & \textbf{15.38} & \textbf{9.09} & \textbf{0.00} & 0.00 & 18.18 & 5.61 \\
& Human Data   & 6.67  & 0.00 & 0.00 & \textbf{7.41} & \textbf{4.29} & 7.69 & \textbf{9.09} & \textbf{0.00} & \textbf{14.29} & 0.00 & 4.67 \\
\cmidrule(lr){2-13}
& \ours   & \textbf{10.00} & \textbf{25.00} & \textbf{3.45} & 3.70 & \textbf{4.29} & \textbf{15.38} & \textbf{9.09} & \textbf{0.00} & \textbf{14.29} & \textbf{27.27} & \textbf{7.94} \\
\midrule

\multirow{4}{*}{Qwen3-VL-8B}
& Zero-Shot    & 23.33 & \textbf{37.50} & 6.90 & 14.81 & \textbf{7.14} & 30.77 & 27.27 & 25.00 & 42.85 & 27.27 & 16.82 \\
& Task-Driven  & 23.33 & 25.00 & 10.34 & \textbf{18.52} & 5.71 & \textbf{38.46} & \textbf{45.45} & 25.00 & 42.85 & 27.27 & 17.75 \\
& Human Data   & 13.33 & \textbf{37.50} & \textbf{13.79} & 14.81 & 5.71 & 23.07 & 36.37 & 25.00 & 42.85 & \textbf{36.36} & 16.35 \\
\cmidrule(lr){2-13}
& \ours   & \textbf{26.67} & \textbf{37.50} & \textbf{13.79} & \textbf{18.52} & \textbf{7.14} & \textbf{38.46} & 36.37 & \textbf{50.00} & \textbf{57.14} & 18.18 & \textbf{20.56} \\
\bottomrule
\end{tabular}
}
\caption{Experiment on OSWorld across different applications.}
\label{tab:osworld_results}
\vspace{-0.75em}
\end{table*}

\paragraph{Human Verification of Synthetic Trajectory.}
Our pipeline relies on an LLM-based task verifier to automatically filter synthetic trajectories by checking whether the final GUI state satisfies the summarized task description. To assess the reliability of this verifier—and to validate the trustworthiness of the synthetic trajectories admitted by this automatic filter—we conduct a targeted human audit. We uniformly sample 100 trajectories after verification and ask annotators to independently judge success or failure by inspecting the trajectories and the corresponding task instruction, without access to the verifier’s prediction. The verifier agrees with human judgments on 87 out of 100 cases, yielding an accuracy of 87\%. The remaining 13 disagreements are primarily attributable to minor, borderline cases. Under this sample size, the estimated 95\% confidence interval is 79.0\% to 92.2\%, indicating that the verifier is a reasonably reliable filter while remaining imperfect.
\section{Experiment Setup}

\subsection{Evaluation Benchmarks}
We evaluate our method on computer-use agent benchmarks (1) OSWorld~\cite{xie2024osworld},where each task consists of a natural-language instruction paired with a reproducible desktop VM snapshot (apps/files/window layout) and an execution-based checker that inspects the final state for success. and (2) WindowsAgentArena~\cite{bonatti2024windows}, which is configured with real Windows 11 VMs. Each task restores a fixed snapshot and uses a deterministic Python evaluator that returns a binary success flag.
For both benchmarks, we evaluate only on tasks that are never used for branching during data generation, and we filter out tasks that are judged incompletable by human inspection~\cite{abhyankar2025osworldhuman}. The details of the benchmarks are included in \autoref{app:benchmarks}.

\subsection{Baselines}
To assess the robustness of our synthetic data generation pipeline, we compare against the following baselines:
(1) \textbf{Zero-Shot.} We evaluate the base model without any supervised fine-tuning (SFT). The agent is prompted to produce GUI actions under the standard OSWorld interaction protocol.
(2) \textbf{Task-Driven.} We implement a representative goal-conditioned data synthesis pipeline commonly used in prior work~\cite{sun2025genesis, lai2024autowebglmlargelanguagemodelbased}. Concretely, we provide an initial screenshot of the target application to GPT-5.1 to generate a natural-language task instruction. We then use Claude 4.5 to execute the task in the environment and record the resulting trajectory.
(3) \textbf{Human Data.} We use the public AgentNet~\cite{wangopencua} dataset, which contains human-labeled GUI actions paired with synthetic reasoning, as a human-data baseline. To ensure platform and application consistency, we restrict AgentNet to the subset that matches the evaluation setting (i.e., the same OS/platform and the same target applications as the benchmarks). To ensure a fair comparison, we train all methods with identical training budgets on each benchmark; details are provided in \autoref{app:training_details}.

\subsection{Training Setup}\label{sec:train_setup}
We adopt trajectory-based supervised fine-tuning (SFT) similar to OS-Genesis~\cite{sun2025genesis}, using a
tool-call GUI action space.
At each step $t\!\in\!\mathcal{T}$, the model conditions on the current screenshot $s_t$, two preceding
screenshots $\{s_{t-1},s_{t-2}\}$, and the interaction history $c_t$.
We decompose supervision into (i) predicting a step-level reasoning state $r_t$ and the next tool-call action $a_t$ (ii) predicting only the
next tool-call action $a_t$:
\[
\mathcal{L}_{\text{plan}}
= -\sum_{t\in\mathcal{T}}\log p_\theta\!\left(r_t,a_t \mid s_{t-2},s_{t-1},s_t,c_t\right),
\]
\[
\mathcal{L}_{\text{act}}
= -\sum_{t\in\mathcal{T}}\log p_\theta\!\left(a_t \mid s_{t-2},s_{t-1},s_t,c_t,r_t\right).
\]
The final objective is $\mathcal{L}=\mathcal{L}_{\text{plan}}+\mathcal{L}_{\text{act}}$. To train our agents, we fully fine-tune three vision-language backbones:
\textbf{GLM-4.1V-9B-Base}~\cite{hong2025glm41v}, a general-purpose multimodal backbone with strong screen-text understanding and visual grounding;
\textbf{Qwen2.5-VL-7B-Instruct}~\cite{bai2025qwen25vl}, an instruction-tuned VLM that provides reliable UI-oriented following of natural-language goals under smaller-scale capacity;
and \textbf{Qwen3-VL-8B-Instruct}~\cite{bai2025qwen3vl}, a newer instruction-tuned VLM with improved multimodal reasoning and fine-grained grounding, which we find beneficial for complex GUI steps. All models are trained via full-parameter VLM fine-tuning on 4 × NVIDIA H200 GPUs.

\section{Experiment Results}
\subsection{Main Results}

\paragraph{OSWorld.}
\autoref{tab:osworld_results} reports success rates on OSWorld broken down by application. Across all three backbones, fine-tuning on trajectories generated by our branching pipeline yields the strongest overall performance. In particular, our method improves GLM-4.1V-9B from near-zero zero-shot performance (0.47 overall) to 7.01, indicating that the synthesized trajectories provide essential supervision for models with weaker UI priors. For Qwen2.5-VL-7B, our method reaches 7.94 overall, outperforming both task-driven synthesis (5.61) and human demonstrations from AgentNet (4.67) under matched platform and application settings. Even for the strongest backbone, Qwen3-VL-8B, our data provides a clear lift from 16.82 to 20.56 overall. The degradation on VS Code may be driven by evaluation variance: the VS Code subset includes only 11 tasks. As a result, small changes in outcomes can produce large swings in the reported score in that domain.

\begin{table}[!t]
\centering
\small
\begin{tabular}{llc}
\toprule
\bfseries Model & \bfseries Setting & \bfseries Full Task SR (\%) \\
\midrule
GLM4.1V-9B      & Zero-Shot & \hphantom{0}\num{5.49} \\
                & Task-Driven   & \num{13.19} \\
                & \ours      & \bfseries\num{16.30} \\
\midrule
Qwen2.5-VL-7B   & Zero-Shot & \hphantom{0}\num{4.39} \\
                & Task-Driven   & \num{14.10} \\
                & \ours       & \bfseries\num{15.22} \\
\midrule
Qwen3-VL-8B     & Zero-Shot & \num{23.07}  \\
                & Task-Driven   & \num{27.47} \\
                & \ours  & \bfseries\num{30.76} \\
\bottomrule
\end{tabular}
\caption{Success rate on WindowsAgentArena}
\label{tab:windowsagentarena_results}
\end{table}

\paragraph{WindowsAgentArena.}
\autoref{tab:windowsagentarena_results} shows full-task success rate on WindowsAgentArena. Our branching-generated supervision consistently improves over both zero-shot and the task-driven synthesis baseline across all backbones. For GLM-4.1V-9B, we improve from 5.49 and 13.19 (task-driven) to 16.30. For Qwen2.5-VL-7B, our method increases success from 4.39 and 14.10 to 15.22. For Qwen3-VL-8B, we raise performance from 23.07 and 27.47 to 30.76. These gains suggest that our pipeline transfers cleanly to the Windows environment and provides complementary supervision beyond goal-conditioned task-driven synthesis, capturing OS-specific UI conventions and multi-step workflows in a form that is effective for supervised fine-tuning.

\subsection{Qualitative Analysis}
To better understand the performance gap between \ours and alternative data-generation pipelines, we conduct a qualitative analysis on a representative deep multistep task: "Enable auto-save every 3 min, so that I don't need to hit 'ctrl-s' that much." This task requires navigating complex nested menus. As shown in \autoref{app:qualitative}, the models trained on data from the Task-Driven generation pipeline correctly identifies the intent but hallucinates the UI structure, opening an unrelated panel in Step 3 and subsequently failing to recover. Similarly, the model trained purely on Human Data fails to generalize to the specific phrasing of the command; it enters the wrong area and oscillates between tabs in the wrong interface, unable to progress toward the specific "auto-save" toggle. In contrast, \ours successfully grounds the instruction to the correct path in the options tree, navigating to \texttt{Load/Save $\rightarrow$ General} and setting \texttt{Save AutoRecovery information every} to \texttt{3} minutes before confirming the change.
We attribute this advantage to our branch-point expansion: we deliberately branch from visually rich GUI states such as the LibreOffice \textit{Options} dialog. Because the \texttt{Tools $\rightarrow$ Options} menu centralizes a wide range of configurable settings, reaching this hub state enables many distinct downstream functionalities to be completed by simply selecting different tree nodes and toggling corresponding controls. As a result, our branching produces a diverse family of descendant trajectories that share the same high-level entry path but cover many different settings panels and interactions.



\begin{figure}[t]
\centering
\begin{tikzpicture}
\begin{axis}[
  trim axis left,
  trim axis right,
  outer sep=0pt,
  width=\columnwidth,
  height=0.85\columnwidth,
  ylabel={Success Rate (\%)},
  xtick      ={3,6,10},
  xticklabels={0.3K,0.6K,1K},
  axis x line*=bottom,
  axis y line*=left,
  xmin=0, xmax=12,
  ymin=0, ymax=30,
  ymajorgrids,
  grid style={dotted,gray!45},
  tick label style={font=\small},
  label style={font=\small},
  every axis plot/.append style={very thick},
  nodes near coords,
  every node near coord/.style={font=\scriptsize,yshift=2pt,text=black},
  legend pos=north west,
  legend style={font=\scriptsize, draw=none, nodes={inner sep=1pt}},
  xlabel={Training data size},
  xlabel style={font=\small, yshift=2pt},
]

\addplot+[
  Blue20, mark=*, mark options={fill=Blue20},
  point meta=explicit symbolic
]
coordinates {
  (0,0.93)  [{}]   
  (3,5.14) [5.14]
  (6,7.14)  [7.14]
  (10,7.94) [7.94]
};
\addlegendentry{Qwen2.5-VL-7B}

\node[font=\scriptsize, anchor=south west, xshift=2pt, yshift=2pt] at (axis cs:0,0.93) {0.93};

\addplot+[
  DeepBlue, dashed, mark=*, mark options={fill=DeepBlue},
  point meta=explicit symbolic
]
coordinates {
  (0,16.82)  [{}]   
  (3,13.08) [13.08]
  (6,17.29)  [17.29]
  (10,20.56) [20.56]
};
\addlegendentry{Qwen3-VL-8B}

\node[font=\scriptsize, anchor=west, xshift=2pt] at (axis cs:0,16.82) {16.82};

\addplot+[
  black, dotted, mark=square*, mark options={fill=black},
  point meta=explicit symbolic
]
coordinates {
  (0,0.47)  [{}]   
  (3,2.8) [2.8]
  (6,4.67) [4.67]
  (10,7.01) [7.01]
};
\addlegendentry{GLM-4.1V-9B}

\node[font=\scriptsize, anchor=west, xshift=2pt, yshift=1pt] at (axis cs:0,0.47) {0.47};

\end{axis}
\end{tikzpicture}
\caption{Scaling curve on OSWorld for \textit{In-domain} data.}
\label{fig:scale_osworld}
\end{figure}
\begin{figure}[t]
\centering
\begin{tikzpicture}
\begin{axis}[
  width=\columnwidth,
  height=0.85\columnwidth,
  ylabel={Success Rate (\%)},
  xtick      ={3,6,10,16},
  xticklabels={0.3K,0.6K,1K,1.6K},
  axis x line*=bottom,
  axis y line*=left,
  xmin=0, xmax=16,
  ymin=0, ymax=17,
  ymajorgrids,
  grid style={dotted,gray!45},
  tick label style={font=\small},
  label style={font=\small},
  every axis plot/.append style={very thick},
  nodes near coords,
  every node near coord/.style={font=\scriptsize,yshift=2pt,text=black},
  legend pos=north west,
  legend style={font=\scriptsize, draw=none, nodes={scale=1.2, transform shape}},
xlabel={Training data size},
  xlabel style={font=\small, yshift=2pt},
]

\addplot+[
  Blue20, mark=*, mark options={fill=Blue20},
  point meta=explicit symbolic,
  nodes near coords,
  nodes near coords style={
    font=\small,   
    yshift=2pt,
    text=black
  }
]
coordinates {
  (0,0.93)  [{}]   
  (3,5.14) [5.14]
  (6,7.14)  [7.14]
  (10,7.94) [7.94]
};
\addlegendentry{Ubuntu}

\node[font=\small, anchor=west, xshift=2pt] at (axis cs:0,0.93) {0.93};

\addplot+[
  DeepBlue, dashed, mark=*, mark options={fill=DeepBlue},
  point meta=explicit symbolic,
  nodes near coords,
  nodes near coords style={
    font=\small,
    anchor=north west,
    xshift=3pt,
    yshift=0pt
  },
  visualization depends on={x \as \thisx},
  every node near coord/.append style={
    /utils/exec={\ifdim\thisx pt=16pt\relax
      \tikzset{anchor=north, xshift=0pt, yshift=-2pt}
    \fi}
  }
]
coordinates {
  (0,0.93)  [{}]   
  (3,2.80) [2.80]
  (6,6.07)  [6.07]
  (10,7.47) [7.47]
  (16, 9.95) [9.95]
};
\addlegendentry{Ubuntu\&Win}

\end{axis}
\end{tikzpicture}
\caption{Scaling curve on OSWorld for \textit{Cross-domain} data. }
\label{fig:scale_osworld_qwen2-5}
\end{figure}
\subsection{Data Scaling Analysis}
We study how the amount of synthetic supervision affects computer-use performance on OSWorld, under two data regimes. \emph{In-domain scaling} increases the number of Ubuntu OSWorld trajectories used for SFT. \emph{Cross-domain augmentation} adds trajectories collected on WindowsAgentArena on top of the Ubuntu set, introducing platform and UI conventions that are absent in OSWorld.

\paragraph{Increasing in-domain data yields consistent gains for all three backbone models.} As shown in \autoref{fig:scale_osworld}, for Qwen2.5-VL-7B, success rate increases from 0.93 to 5.14 with 0.3K trajectories and further to 7.94 with 1K trajectories, indicating that even a few hundred trajectories provide a strong supervision signal. GLM-4.1V-9B follows a similar trend, improving from 0.47 to 7.01 when scaling to 1K trajectories. Qwen3-VL-8B benefits most at larger scales, reaching 20.56 at 1K, but exhibits a small-scale dip from 16.82 to 13.08 at 0.3K before recovering. This non-monotonic behavior is consistent with a format mismatch at low data scale, where the model is sensitive to discrepancies between its native action and reasoning template and the action-space and reasoning style used in our collected trajectories. As the dataset grows, the fine-tuning signal becomes more stable and the model learns to consistently map screenshots to the intended tool calls, yielding a clear upward trend.

\paragraph{Cross-domain data can improve performance rather than causing negative transfer.} As shown in \autoref{fig:scale_osworld_qwen2-5}, for Qwen2.5-VL-7B, augmenting the 1K Ubuntu set with 600 WindowsAgentArena trajectories increases success from 7.94 to 9.95. While the mixed Ubuntu and Windows setting is slightly behind the Ubuntu-only curve at smaller scales, the advantage emerges at larger scale, suggesting that additional UI diversity helps the agent learn more robust grounding and recovery behaviors. Importantly, these gains come from relatively small amounts of data because our trajectories are high-signal. We generate trajectories from a curated set of seed tasks, expand them at branch points with substantial UI state changes, and apply step-level verification and filtering to remove inconsistent tool calls and low-quality segments. The resulting supervision is dense and coherent at the step level, with fewer off-trajectory hallucinations, which helps explain why performance improves sharply already at 0.3K and continues to scale with additional data.

\subsection{Ablations on \ours Components}

We evaluate the contribution of our step-level quality controls by constructing a training set that removes both (i) prefix step filtering and (ii) post-branch denoising. Concretely, rather than sampling multiple candidate rationales and retaining only the one that best matches the observed transition, we generate a single rationale for every recorded step by prompting the model with the pre-action state, the executed action, and the post-action state, regardless of whether the step is noisy or not. In addition, we keep all post-branch steps without applying our intention-consistency checking.

We then fine-tune the same agent backbones on this unfiltered set of data. As shown in \autoref{tab:step-level-effect}, eliminating step-level filtering and post-branch denoising consistently degrades performance across different models. This consistent decline indicates that enforcing step-level filtering and denoising are important for effective supervision.

\begin{table}[!t]
    \centering
    \small
    \begin{tabular}{lcc}
        \toprule
        \textbf{Model} & \textbf{w/ filt+denoise} & \textbf{w/o (unfiltered)} \\
        \midrule
        \textbf{Qwen3-VL-8B}   & 20.56 & 19.15 \\
        \textbf{Qwen2.5-VL-7B} & 7.94  & 7.01 \\
        \textbf{GLM4.1V-9B} & 7.01 & 6.54 \\
        \bottomrule
    \end{tabular}
    \caption{Effect of step-level filtering and post-branch denoising on OSWorld success rate (\%).}
    \label{tab:step-level-effect}
\end{table}


\section{Conclusion}
We introduced a trajectory expansion framework for scalable desktop GUI data generation. By branching from verified seed demonstrations at meaningful state-change points, our pipeline produces diverse, executable trajectories with strong task grounding, supported by task-conditioned filtering and post-branch denoising. Across two desktop benchmarks and multiple VLM backbones, fine-tuning on our data consistently outperforms baselines and improves cross-application and cross-platform generalization. These results highlight that anchoring expansion to verified prefixes can yield reliable long-horizon supervision while reducing sensitivity to imperfect rollout policies. More broadly, our findings suggest that state-grounded branching offers a practical path to scaling high-quality GUI supervision without requiring flawless exploration or exhaustive human demonstrations. Future work includes stronger verification, improved branching policies, and scaling to additional environments and interaction modalities.

\section*{Limitations}

We evaluate on desktop computer-use benchmarks (OSWorld, WindowsAgentArena), which remain comparatively under-explored for end-to-end GUI agents and align well with our environment setup.
While our experiments focus on desktop workflows, the proposed branch-point expansion pipeline is not tied to a specific platform. We encourage future researchers to extend our work to mobile and web settings to study performance under different UI primitives and accessibility signals.

\bibliography{custom}

\clearpage
\appendix

\section{Details of Benchmarks}

\label{app:benchmarks}

\subsection{OSWorld}

OSWorld~\citep{xie2024osworld} is a real-computer benchmark that wraps full desktop operating systems (primarily Ubuntu) in a controlled virtual machine (VM) environment. Each task is specified by (i) a natural-language instruction describing the user-level goal, (ii) a reproducible VM snapshot encoding the initial desktop state (open applications, files, and window layout), and (iii) an execution script that inspects the final VM state and returns a scalar reward. In our experiments, we follow the benchmark’s execution-based protocol and treat a task as successful if and only if the checker returns a positive signal. The agent interacts with OSWorld exclusively through GUI-level actions: it receives screenshots (and optional structured metadata such as accessibility trees) and issues mouse movements, clicks, scrolls, keyboard events, and window-management operations. We use the standard OSWorld action interface without modifying the underlying environment. Episodes terminate either when the benchmark signals task completion, when the environment reports a failure (e.g., application crash), or when a fixed step budget is reached. We always evaluate on tasks that are disjoint from those used as gold trajectories for data generation.

\subsection{WindowsAgentArena}

WindowsAgentArena~\citep{bonatti2024windows} builds on the OSWorld framework to provide an execution-based benchmark focused exclusively on the Windows operating system. Tasks are defined on real Windows 11 VMs packaged inside Docker containers. Each task includes a configuration that restores a Windows snapshot (installed applications, files, and initial foreground windows) and a deterministic Python-based evaluator that inspects the final state and returns a binary success flag. The observation and action spaces mirror real Windows usage: the agent receives the current screen (and optionally accessibility information or structured overlays) and produces GUI actions such as clicks, key presses, and scrolls. The benchmark covers representative Windows workloads, including document and spreadsheet editing, web browsing, system and file-management operations, coding in Visual Studio Code, and basic utilities (e.g., Notepad, Paint, Clock, Settings). WindowsAgentArena is designed to scale out over Azure Machine Learning: tasks are automatically sharded across multiple Windows VMs, and their results are aggregated after all episodes finish. In our experiments, we evaluate agents using the official deterministic evaluation scripts and report success rate over the full task set, again ensuring that all evaluation tasks are disjoint from any tasks used for training or data generation.

\section{Experimental Details}
\label{app:exp_details}

\subsection{Desktop action space and tool schema}
All trajectories are represented as calls to a single function tool,
\texttt{computer\_use}, which matches the inference-time interface used by our
agent implementation. Each step predicts exactly one tool call with an
\texttt{action} field and optional arguments (coordinates, text, keys, etc.).
\autoref{tab:action_space_desktop} lists the complete action space used in our
training code.

\begin{table}[h]
\centering
\resizebox{0.9\columnwidth}{!}{%
\begin{tabular}{@{}ll@{}}
\toprule
Action & Description \\
\midrule
\texttt{mouse\_move}      & Move the cursor to a target \texttt{coordinate}. \\
\texttt{left\_click}      & Left click at a target \texttt{coordinate}. \\
\texttt{right\_click}     & Right click at a target \texttt{coordinate}. \\
\texttt{middle\_click}    & Middle click at a target \texttt{coordinate}. \\
\texttt{double\_click}    & Double click at a target \texttt{coordinate}. \\
\texttt{left\_click\_drag}& Drag from \texttt{start\_coordinate} to \texttt{coordinate}. \\
\texttt{scroll}           & Scroll vertically by \texttt{pixels} (sign indicates direction). \\
\texttt{type}             & Type a string into the focused input (\texttt{text}). \\
\texttt{key}              & Press a key or key-combo (\texttt{keys}). \\
\texttt{wait}             & Wait for UI to update (\texttt{time} in seconds). \\
\midrule
\texttt{terminate}        & End the task with \texttt{status}\,$\in\{\texttt{success},\texttt{failure}\}$. \\
\bottomrule
\end{tabular}
}
\caption{Action space for our GUI setting}
\label{tab:action_space_desktop}
\end{table}

\subsection{Prompt}
The prompt for identify branch state, branch task generation, execution, summarizing and verification, and reasoning denoising and filtering are shown in \autoref{fig:branch-point-prompt}, \autoref{fig:branch-task-gen-prompts}, \autoref{fig:desktop-train-prompt}, \autoref{fig:task-gen-verify-prompts}, and \autoref{fig:reasoning-prompts} respectively.
\begin{figure*}[t]
    \centering
    \setlength{\fboxrule}{0.85pt}
    \fbox{ \footnotesize
        \parbox{1.0\textwidth}{\texttt{\textbf{Prompt for Branch-Point Identification }\\
You are an expert in analyzing GUI automation trajectories. I will provide you with an original task description, a trajectory summary, and \textbf{screenshots for multiple steps} in the trajectory (ordered by time).\\
Your job is to identify \textbf{optimal branch states} along this successful trajectory: intermediate states \emph{after certain steps} from which we can propose \textbf{new, different task descriptions} that start from that state.\\
\\
\textbf{Original task description:} \{task\_description\}\\
\textbf{Trajectory summary (\{num\_steps\} steps):} \{trajectory\_summary\}\\
\textbf{Screenshots (ordered states):} \texttt{<image>} \ldots \texttt{<image>}\\
\\
\textbf{Guidelines for good branch points:}\\
1) Use the screenshots to locate \textbf{visually rich, decision-heavy states} (multiple actionable UI elements).\\
2) Prefer states that \textbf{unlock many possible next tasks} (post-navigation, post-opening, post-login, post-setup).\\
3) Prefer states that \textbf{Show new information that is previously unknown} (scroll down to see new images.)
4) Avoid states where the next action is essentially forced by the original task (over-specialized).\\
5) Aim for \textbf{3--5} high-quality branch points.\\
6) For each branch point, provide 2--3 examples of \textbf{new tasks} that could reasonably start from that state.\\
\\
\textbf{Output (JSON only):}\\
\texttt{\{\\
\ \ "branch\_points": [\\
\ \ \ \ \{\\
\ \ \ \ \ \ "after\_step": <step\_number>,\\
\ \ \ \ \ \ "reason": "<why this state is a good starting point for new tasks>",\\
\ \ \ \ \ \ "new\_task\_examples": "<2--3 concrete new tasks doable from this state>"\\
\ \ \ \ \}\\
\ \ ]\\
\}}\\
Return \textbf{only} valid JSON. No extra text, no code fences.
        }}
    }

    \captionsetup{labelformat=default, name=Prompt}
    \caption{Prompts used to identify branch states for task diversification. }
    \label{fig:branch-point-prompt}
\end{figure*}

\begin{figure*}[t]
    \centering
    \setlength{\fboxrule}{0.85pt}

    \fbox{ \footnotesize
        \parbox{1.0\textwidth}{\texttt{\textbf{Prompt for Progress Summary at Branch State}\\
You are an analyst. Given the original task, prior steps (action + reasoning), and the current UI state image, write a \textbf{chronological} summary (within \textbf{3 sentences}) describing what the agent has done so far up to this point.\\
Return \textbf{only} the summary text. The summary should focus on what the agent has done on the UI (navigation, opened panes, edited content), not on abstract plans.\\
\\
\textbf{Original task instruction:} \{original\_instruction\}\\
\textbf{Branch point:} after step \{after\_step\}\\
\textbf{Prior steps (action + reasoning):} \{steps\_digest\}\\
\textbf{Current UI state screenshot:} \texttt{<image>}\\
\\
Return \textbf{only} the summary text (no JSON, no bullet points, no extra formatting).
        }}
    }

    \vspace{0.6em}

    \fbox{ \footnotesize
        \parbox{1.0\textwidth}{\texttt{\textbf{Prompt for Branch Task Generation}\\
You propose natural, feasible follow-up tasks based on the \textbf{current UI state}.\\
When generating a task, describe it as a \textbf{complete end-to-end objective} an expert could accomplish from start to finish, not merely a single interaction at the current state.\\
\\
\textbf{Requirements for each proposed task:}\\
- Be specific and clearly defined, but do \textbf{not} enumerate step-by-step procedures.\\
- Must have \textbf{verifiable success criteria}.\\
- Should feel like a \textbf{natural extension} of what has been done so far.\\
- Should be completable within \textbf{5--15 steps}.\\
- Must be \textbf{different} from the original task.\\
- Must \textbf{not} require authentication or login.\\
- Avoid proposing tasks that are redundant with or too similar to the \textbf{previously proposed tasks} list.\\
\\
\textbf{Agent progress summary:} \{progress\_summary\}\\
\textbf{Original task instruction:} \{original\_instruction\}\\
\textbf{Branch-point reason (optional context):} \{branch\_reason\}\\
\textbf{Example new tasks from branch analysis (optional):} \{branch\_task\_examples\}\\
\textbf{Previously proposed tasks (avoid duplicates):} \{previous\_tasks\}\\
\textbf{Current UI state screenshot:} \texttt{<image>}\\
\\
\textbf{Your task:} Generate \{num\_tasks\} tasks that are mutually different and also different from the previously proposed tasks.\\
Return \textbf{only} valid JSON of the form:\\
\texttt{\{\\
\ \ "tasks": [\\
\ \ \ \ \{"description": "General task description"\},\\
\ \ \ \ \ldots\\
\ \ ]\\
\}}\\
No extra text, no code fences.
        }}
    }

    \captionsetup{labelformat=default, name=Prompt}
    \caption{Prompts used for branch trajectory generation: summarizing progress at a branch state (top) and generating diverse follow-up tasks conditioned on the current UI state (bottom).}
    \label{fig:branch-task-gen-prompts}
\end{figure*}
\begin{figure*}[t]
    \centering
    \setlength{\fboxrule}{0.85pt}

    \fbox{ \footnotesize
        \parbox{1.0\textwidth}{\texttt{\textbf{Prompt for Planning-Oriented Supervision (Reasoning + Tool Call)}\\
You are an agent operating a desktop GUI via a function tool call interface. I will provide you with
a task instruction, a short summary of previous actions, and up to three screenshots (two past screenshots and the current one). \\
\textbf{Instruction:} \{task\_instruction\} \\
\textbf{Previous actions:} \{previous\_actions\} \\
Please generate the next move based on the current UI state. Your response must follow the exact format below:\\
(1) \textbf{Reasoning:} one short sentence describing the intended UI action.\\
(2) a single \textbf{\texttt{<tool\_call>}} block containing only JSON of the form
\texttt{\{"name":"computer\_use","arguments":\{...\}\}}.
        }}
    }

    \vspace{0.6em}

    \fbox{ \footnotesize
        \parbox{1.0\textwidth}{\texttt{\textbf{Prompt for Action-Oriented Supervision (Tool Call Only)}\\
You are an agent operating a desktop GUI via a function tool call interface. I will provide you with
a task instruction, a short summary of previous actions, up to three screenshots, and the next-step reasoning. \\
\textbf{Instruction:} \{task\_instruction\} \\
\textbf{Previous actions:} \{previous\_actions\} \\
\textbf{Reasoning:} \{next\_step\_reasoning\} \\
Please generate only the tool call for the next step, as a single \textbf{\texttt{<tool\_call>}} block with JSON
\texttt{\{"name":"computer\_use","arguments":\{...\}\}}.
        }}
    }

    \captionsetup{labelformat=default, name=Prompt}
    \caption{Prompts used to train our desktop agents. Planning-Oriented Supervision predicts both reasoning and the tool call; Action-Oriented Supervision conditions on the reasoning and predicts the tool call only.}
    \label{fig:desktop-train-prompt}
\end{figure*}

\begin{figure*}[t]
    \centering
    \setlength{\fboxrule}{0.85pt}

    \fbox{ \footnotesize
        \parbox{1.0\textwidth}{\texttt{\textbf{Prompt for Candidate Reasoning Generation}\\
You are helping backfill missing reasoning for \textbf{replay (pre-branch)} steps in a GUI trajectory. You will be given the task description, a summary of prior replay actions, and replay screenshots up to the current state.\\
Your job is to propose multiple \emph{distinct} plausible \textbf{single-step} next actions that could have led to the next recorded state, each paired with a short first-person reasoning.\\
\textbf{Task description:} \{task\_description\}\\
\textbf{History of completed replay steps:} \{history\_summary\}\\
\textbf{Screenshots (replay history + current state):} \texttt{<image>} \ldots \texttt{<image>}\\
Generate \{num\_candidates\} candidates. Each \textbf{action\_summary} must be a \textbf{single} concrete UI action (click / type / select / copy / open), not a compound action.\\
Each \textbf{reasoning} must be 1--2 sentences in \textbf{first-person} describing what is visible and why this action is the next step.\\
Return \textbf{only} valid JSON of the form: \texttt{\{"candidates":[\{"action\_summary":"...","reasoning":"..."\}, ...]\}}.
        }}
    }

    \vspace{0.6em}

    \fbox{ \footnotesize
        \parbox{1.0\textwidth}{\texttt{\textbf{Prompt for Candidate Verification}\\
You are verifying whether a proposed natural-language candidate action matches the \textbf{recorded replay action} for this step.\\
Use the recorded action script and the visual change between the before/after screenshots to decide if they refer to the same UI operation.\\
\textbf{Recorded replay action script:} \{action\_script\}\\
\textbf{Candidate action:} \{candidate\_action\}\\
\textbf{Screenshots:} \texttt{<image>} (before) \;\; \texttt{<image>} (after)\\
Answer \textbf{only} with JSON: \texttt{\{"match": true|false, "explanation": "..."\}}.\\
Set \textbf{match=true} only if the candidate accurately describes the observed UI change and aligns with the recorded script.
        }}
    }

    \vspace{0.6em}
    \captionsetup{labelformat=default, name=Prompt}
    \caption{Prompts for Candidate Reasoning Generation and filtering}
    \label{fig:reasoning-prompts}
\end{figure*}

\begin{figure*}[t]
    \centering
    \setlength{\fboxrule}{0.85pt}

    \fbox{ \footnotesize
        \parbox{1.0\textwidth}{\texttt{\textbf{Prompt for Unified Task Description Generation}\\
You are a task generation expert. Given a list of actions and screenshots, produce a \textbf{single concise task description} that would be accomplished by performing these actions in order on the computer.\\
Guidelines: (1) The task must be clear, specific, and feasible. (2) Include all necessary information, and do not require unspecified external context. (3) Describe the \emph{overall goal}, not step-by-step low-level operations. (4) If the provided task description matches the actions and screenshot progression, you may return it unchanged; otherwise, rewrite a correct one.\\
Examples: ``Increase the brightness and contrast of the image on Slide 2 to make its details more visible.'';
``Calculate and display the maximum Revenue value in a new cell below the data.'';\\
``Set the Word Wrap Column value to 120 characters to allow longer lines.'';\\
``Configure the browser to open a specific website as the homepage and startup page.''\\
\textbf{App/domain:} \{app\_name\}\\
\textbf{List of actions:} \{action\_summary\}\\
\textbf{Provided task description (if any):} \{provided\_task\_descriptions\}\\
\textbf{Screenshots in chronological order:} \texttt{<image>} \ldots \texttt{<image>}\\
Your task: If the provided description matches the full action sequence and screenshot progression, return it. Otherwise, propose a single task description that is accomplished by executing the actions in the given order.\\
Return \textbf{only} the task description text (no bullets, no quotes, no extra explanation).
        }}
    }

    \vspace{0.6em}

    \fbox{ \footnotesize
        \parbox{1.0\textwidth}{\texttt{\textbf{Prompt for Task Success Verification}\\
You are a precise evaluator. Given a task description, the full action trajectory, and screenshots, determine whether the task was successfully completed.\\
\textbf{Task description:} \{task\_description\}\\
\textbf{Action trajectory:} \{action\_summary\}\\
\textbf{Screenshots (chronological):} \texttt{<image>} \ldots \texttt{<image>}\\
Instruction: Assess whether the actions and the final UI states shown in the screenshots complete the task.\\
Respond in \textbf{strict JSON only} (no code fences, no extra text) with exactly this structure:\\
\texttt{\{}\\
\texttt{\ \ "success": true|false,}\\
\texttt{\ \ "explanation": "short explanation"}\\
\texttt{\}}\\
Keep the explanation brief and grounded in the visible outcomes.
        }}
    }

    \captionsetup{labelformat=default, name=Prompt}
    \caption{Prompts used to (1) generate a unified task description from an action trace and screenshots, and (2) verify whether the trajectory successfully completes the task.}
    \label{fig:task-gen-verify-prompts}
\end{figure*}

\subsection{Tasks selected for branching}
To avoid training--test leakage, we exclude from evaluation all benchmark tasks whose IDs are used as branching seeds during data generation. \autoref{fig:osworld_branch_task_ids} and \autoref{fig:winarena_branch_task_ids} list the selected seed task IDs for OSWorld and WindowsAgentArena, respectively.

\begin{figure*}[t]
  \centering
  \fbox{%
    \begin{minipage}{0.95\textwidth}
      \captionsetup{type=figure}
      \small
      \textbf{OSWorld branching task IDs}\par
      \vspace{2pt}\hrule\vspace{6pt}
      \ttfamily\tiny
      \setlength{\columnsep}{10pt}
      \begin{multicols}{3}
      \begin{enumerate}[leftmargin=*,label={},itemsep=0pt,topsep=0pt]
        \item 70bca0cc-c117-427e-b0be-4df7299ebeb6
        \item 7efeb4b1-3d19-4762-b163-63328d66303b
        \item 12382c62-0cd1-4bf2-bdc8-1d20bf9b2371
        \item d53ff5ee-3b1a-431e-b2be-30ed2673079b
        \item eb03d19a-b88d-4de4-8a64-ca0ac66f426b
        \item 72b810ef-4156-4d09-8f08-a0cf57e7cefe
        \item a434992a-89df-4577-925c-0c58b747f0f4
        \item 04578141-1d42-4146-b9cf-6fab4ce5fd74
        \item 0d8b7de3-e8de-4d86-b9fd-dd2dce58a217
        \item c82632a4-56b6-4db4-9dd1-3820ee3388e4
        \item af2d657a-e6b3-4c6a-9f67-9e3ed015974c
        \item ce2b64a2-ddc1-4f91-8c7d-a88be7121aac
        \item 276cc624-87ea-4f08-ab93-f770e3790175
        \item 7a4deb26-d57d-4ea9-9a73-630f66a7b568
        \item 66399b0d-8fda-4618-95c4-bfc6191617e9
        \item bb5e4c0d-f964-439c-97b6-bdb9747de3f4
        \item e0df059f-28a6-4169-924f-b9623e7184cc
        \item 8979838c-54a5-4454-a2b8-3d135a1a5c8f
        \item 68a25bd4-59c7-4f4d-975e-da0c8509c848
        \item eabc805a-bfcf-4460-b250-ac92135819f6
        \item 5ced85fc-fa1a-4217-95fd-0fb530545ce2
        \item 3aaa4e37-dc91-482e-99af-132a612d40f3
        \item 51b11269-2ca8-4b2a-9163-f21758420e78
        \item 4bcb1253-a636-4df4-8cb0-a35c04dfef31
        \item 1273e544-688f-496b-8d89-3e0f40aa0606
        \item f23acfd2-c485-4b7c-a1e7-d4303ddfe864
        \item 02ce9a50-7af2-47ed-8596-af0c230501f8
        \item ec4e3f68-9ea4-4c18-a5c9-69f89d1178b3
        \item 73c99fb9-f828-43ce-b87a-01dc07faa224
        \item 3ce045a0-877b-42aa-8d2c-b4a863336ab8
        \item 6ada715d-3aae-4a32-a6a7-429b2e43fb93
        \item 48c46dc7-fe04-4505-ade7-723cba1aa6f6
        \item a01fbce3-2793-461f-ab86-43680ccbae25
        \item f178a4a9-d090-4b56-bc4c-4b72a61a035d
        \item 4e6fcf72-daf3-439f-a232-c434ce416af6
        \item ef9d12bd-bcee-4ba0-a40e-918400f43ddf
        \item 0e763496-b6bb-4508-a427-fad0b6c3e195
        \item a097acff-6266-4291-9fbd-137af7ecd439
        \item 0b17a146-2934-46c7-8727-73ff6b6483e8
        \item c288e301-e626-4b98-a1ab-159dcb162af5
        \item 7dbc52a6-11e0-4c9a-a2cb-1e36cfda80d8
        \item 5d901039-a89c-4bfb-967b-bf66f4df075e
        \item 227d2f97-562b-4ccb-ae47-a5ec9e142fbb
        \item 7a4e4bc8-922c-4c84-865c-25ba34136be1
        \item 5203d847-2572-4150-912a-03f062254390
        \item 4127319a-8b79-4410-b58a-7a151e15f3d7
        \item f3b19d1e-2d48-44e9-b4e1-defcae1a0197
        \item 7a5a7856-f1b6-42a4-ade9-1ca81ca0f263
        \item 2b94c692-6abb-48ae-ab0b-b3e8a19cb340
        \item 1e8df695-bd1b-45b3-b557-e7d599cf7597
        \item ac1b39ff-ee4d-4483-abce-c117e98942f0
        \item 42e0a640-4f19-4b28-973d-729602b5a4a7
        \item d38192b0-17dc-4e1d-99c3-786d0117de77
        \item 3a7c8185-25c1-4941-bd7b-96e823c9f21f
        \item 4c26e3f3-3a14-4d86-b44a-d3cedebbb487
        \item 94d95f96-9699-4208-98ba-3c3119edf9c2
        \item 82279c77-8fc6-46f6-9622-3ba96f61b477
        \item b337d106-053f-4d37-8da0-7f9c4043a66b
        \item 9cf05d24-6bd9-4dae-8967-f67d88f5d38a
        \item f9be0997-4b7c-45c5-b05c-4612b44a6118
        \item 0bf05a7d-b28b-44d2-955a-50b41e24012a
        \item 81c425f5-78f3-4771-afd6-3d2973825947
        \item 77b8ab4d-994f-43ac-8930-8ca087d7c4b4
        \item 59155008-fe71-45ec-8a8f-dc35497b6aa8
        \item 4e60007a-f5be-4bfc-9723-c39affa0a6d3
        \item 9656a811-9b5b-4ddf-99c7-5117bcef0626
        \item 215dfd39-f493-4bc3-a027-8a97d72c61bf
        \item 0512bb38-d531-4acf-9e7e-0add90816068
        \item d16c99dc-2a1e-46f2-b350-d97c86c85c15
        \item da46d875-6b82-4681-9284-653b0c7ae241
        \item 51f5801c-18b3-4f25-b0c3-02f85507a078
        \item 39be0d19-634d-4475-8768-09c130f5425d
        \item 2b9493d7-49b8-493a-a71b-56cd1f4d6908
        \item b148e375-fe0b-4bec-90e7-38632b0d73c2
        \item 2ae9ba84-3a0d-4d4c-8338-3a1478dc5fe3
        \item 3c8f201a-009d-4bbe-8b65-a6f8b35bb57f
        \item 869de13e-bef9-4b91-ba51-f6708c40b096
        \item f723c744-e62c-4ae6-98d1-750d3cd7d79d
        \item 8ba5ae7a-5ae5-4eab-9fcc-5dd4fe3abf89
        \item 26660ad1-6ebb-4f59-8cba-a8432dfe8d38
        \item a10b69e1-6034-4a2b-93e1-571d45194f75
        \item d06f0d4d-2cd5-4ede-8de9-598629438c6e
        \item f4aec372-4fb0-4df5-a52b-79e0e2a5d6ce
        \item 788b3701-3ec9-4b67-b679-418bfa726c22
        \item 21760ecb-8f62-40d2-8d85-0cee5725cb72
        \item 3161d64e-3120-47b4-aaad-6a764a92493b
        \item ecc2413d-8a48-416e-a3a2-d30106ca36cb
        \item 37608790-6147-45d0-9f20-1137bb35703d
        \item 28cc3b7e-b194-4bc9-8353-d04c0f4d56d2
        \item 7b7617bd-57cc-468e-9c91-40c4ec2bcb3d
        \item f7dfbef3-7697-431c-883a-db8583a4e4f9
        \item 7767eef2-56a3-4cea-8c9f-48c070c7d65b
        \item af23762e-2bfd-4a1d-aada-20fa8de9ce07
        \item 58565672-7bfe-48ab-b828-db349231de6b
        \item fba2c100-79e8-42df-ae74-b592418d54f4
        \item 0e5303d4-8820-42f6-b18d-daf7e633de21
        \item c2751594-0cd5-4088-be1b-b5f2f9ec97c4
        \item 7ae48c60-f143-4119-b659-15b8f485eb9a
        \item f3977615-2b45-4ac5-8bba-80c17dbe2a37
        \item 53ad5833-3455-407b-bbc6-45b4c79ab8fb
        \item 386dbd0e-0241-4a0a-b6a2-6704fba26b1c
        \item dfac9ee8-9bc4-4cdc-b465-4a4bfcd2f397
        \item b4f95342-463e-4179-8c3f-193cd7241fb2
        \item 3299584d-8f11-4457-bf4c-ce98f7600250
        \item 9439a27b-18ae-42d8-9778-5f68f891805e
        \item 4de54231-e4b5-49e3-b2ba-61a0bec721c0
        \item a5bbbcd5-b398-4c91-83d4-55e1e31bbb81
        \item 337d318b-aa07-4f4f-b763-89d9a2dd013f
        \item e2b5e914-ffe1-44d2-8e92-58f8c5d92bb2
        \item dd84e895-72fd-4023-a336-97689ded257c
        \item a728a36e-8bf1-4bb6-9a03-ef039a5233f0
        \item 9b7bc335-06b5-4cd3-9119-1a649c478509
        \item a74b607e-6bb5-4ea8-8a7c-5d97c7bbcd2a
        \item ecb0df7a-4e8d-4a03-b162-053391d3afaf
        \item 00fa164e-2612-4439-992e-157d019a8436
        \item ee9a3c83-f437-4879-8918-be5efbb9fac7
        \item f79439ad-3ee8-4f99-a518-0eb60e5652b0
      \end{enumerate}
      \end{multicols}
    \end{minipage}%
  }
  \caption{OSWorld task IDs used as branching seeds.}
  \label{fig:osworld_branch_task_ids}
\end{figure*}

\begin{figure*}[t]
  \centering
  \fbox{%
    \begin{minipage}{0.95\textwidth}
      \captionsetup{type=figure}
      \small
      \textbf{WindowsAgentArena branching task IDs}\par
      \vspace{2pt}\hrule\vspace{6pt}
      \ttfamily\tiny
      \setlength{\columnsep}{10pt}
      \begin{multicols}{3}
      \begin{enumerate}[leftmargin=*,label={},itemsep=0pt,topsep=0pt]
        \item 0810415c-bde4-4443-9047-d5f70165a697-WOS
        \item 0acbd372-ca7a-4507-b949-70673120190f-WOS
        \item 0cecd4f3-74de-457b-ba94-29ad6b5dafb6-WOS
        \item 15f8de6e-3d39-40e4-af17-bdbb2393c0d9-WOS
        \item 1d17d234-e39d-4ed7-b46f-4417922a4e7c-WOS
        \item 1e8df695-bd1b-45b3-b557-e7d599cf7597-WOS
        \item 21ab7b40-77c2-4ae6-8321-e00d3a086c73-WOS
        \item 276cc624-87ea-4f08-ab93-f770e3790175-2-WOS
        \item 39aa4e37-dc91-482e-99af-132a612d40f3-WOS
        \item 3ef2b351-8a84-4ff2-8724-d86eae9b842e-WOS
        \item 44dbac63-32bf-4cd2-81b4-ad6803ec812d-WOS
        \item 4e6fcf72-daf3-439f-a232-c434ce416af6-WOS
        \item 5548314e-d807-4e9e-97e9-b3a4f9fd634f-WOS
        \item 70745df8-f2f5-42bd-8074-fbc10334fcc5-2-WOS
        \item 930fdb3b-11a8-46fe-9bac-577332e2640e-WOS
        \item 936321ce-5236-426a-9a20-e0e3c5dc536f-WOS
        \item INF-5ac2891a-eacd-4954-b339-98abba077adb-WOS
        \item d53ff5ee-3b1a-431e-b2be-30ed2673079b-WOS
        \item e528b65e-1107-4b8c-8988-490e4fece599-WOS
        \item ec71221e-ac43-46f9-89b8-ee7d80f7e1c5-WOS
        \item 1c9d2c6c-ae4b-4359-9a93-9d3c42f48417-wos
        \item 5b46f4a4-1a78-4860-ad92-76e051fa7efc-wos
        \item 98cfcec4-c74e-4faa-b70d-664fb0a1d457-wos
        \item 1376d5e7-deb7-471a-9ecc-c5d4e155b0c8-wos
        \item 58f493b5-5a96-4450-99ca-7cebe144c7e5-wos
        \item b27399ae-e91a-4055-9406-472372e0f5c7-wos
        \item 016c9a9d-f2b9-4428-8fdb-f74f4439ece6-WOS
        \item 01b269ae-2111-4a07-81fd-3fcd711993b0-WOS
        \item 035f41ba-6653-43ab-aa63-c86d449d62e5-WOS
        \item 04d9aeaf-7bed-4024-bedb-e10e6f00eb7f-WOS
        \item 0e47de2a-32e0-456c-a366-8c607ef7a9d2-WOS
        \item 1de60575-bb6e-4c3d-9e6a-2fa699f9f197-WOS
        \item 2d292a2d-686b-4e72-80f7-af6c232b1258-WOS
        \item 30e3e107-1cfb-46ee-a755-2cd080d7ba6a-WOS
        \item 3bad5766-5186-42be-abe1-12eacc798d3a-WOS
        \item 4de54231-e4b5-49e3-b2ba-61a0bec721c0-WOS
        \item 4e60007a-f5be-4bfc-9723-c39affa0a6d3-WOS
        \item 5d353deb-c4b0-4126-a99e-5490817b48cb-WOS
        \item 5f8601f8-6e90-4d2c-91bb-eb5836ad1d5c-WOS
        \item 7a4e4bc8-922c-4c84-865c-25ba34136be1-WOS
        \item 8b1ce5f2-59d2-4dcc-b0b0-666a714b9a14-WOS
        \item 9656a811-9b5b-4ddf-99c7-5117bcef0626-wos
        \item 99146c54-4f37-4ab8-9327-5f3291665e1e-wos
        \item 9ed02102-6b28-4946-8339-c028166e9512-WOS
        \item INF-7aeae0e2-70ee-4705-821d-1bba5d5b2ddd-WOS
        \item INF-847a96b6-df94-4927-97e6-8cc9ea66ced7-WOS
        \item INF-cb130f0d-d36f-4302-9838-b3baf46139b6-WOS
        \item INF-dcbe20e8-647f-4f1d-8696-f1c5bbb570e3-WOS
        \item a5bbbcd5-b398-4c91-83d4-55e1e31bbb81-WOS
        \item e2b5e914-ffe1-44d2-8e92-58f8c5d92bb2-WOS
        \item ea98c5d7-3cf9-4f9b-8ad3-366b58e0fcae-WOS
      \end{enumerate}
      \end{multicols}
    \end{minipage}%
  }
  \caption{WindowsAgentArena task IDs used as branching seeds.}
  \label{fig:winarena_branch_task_ids}
\end{figure*}

\section{Qualitative Analysis Example}
\label{app:qualitative}
\begin{figure*}[t]
\centering
\setlength{\tabcolsep}{2pt}

\begin{subfigure}{0.32\linewidth}
  \centering
  \includegraphics[width=\linewidth]{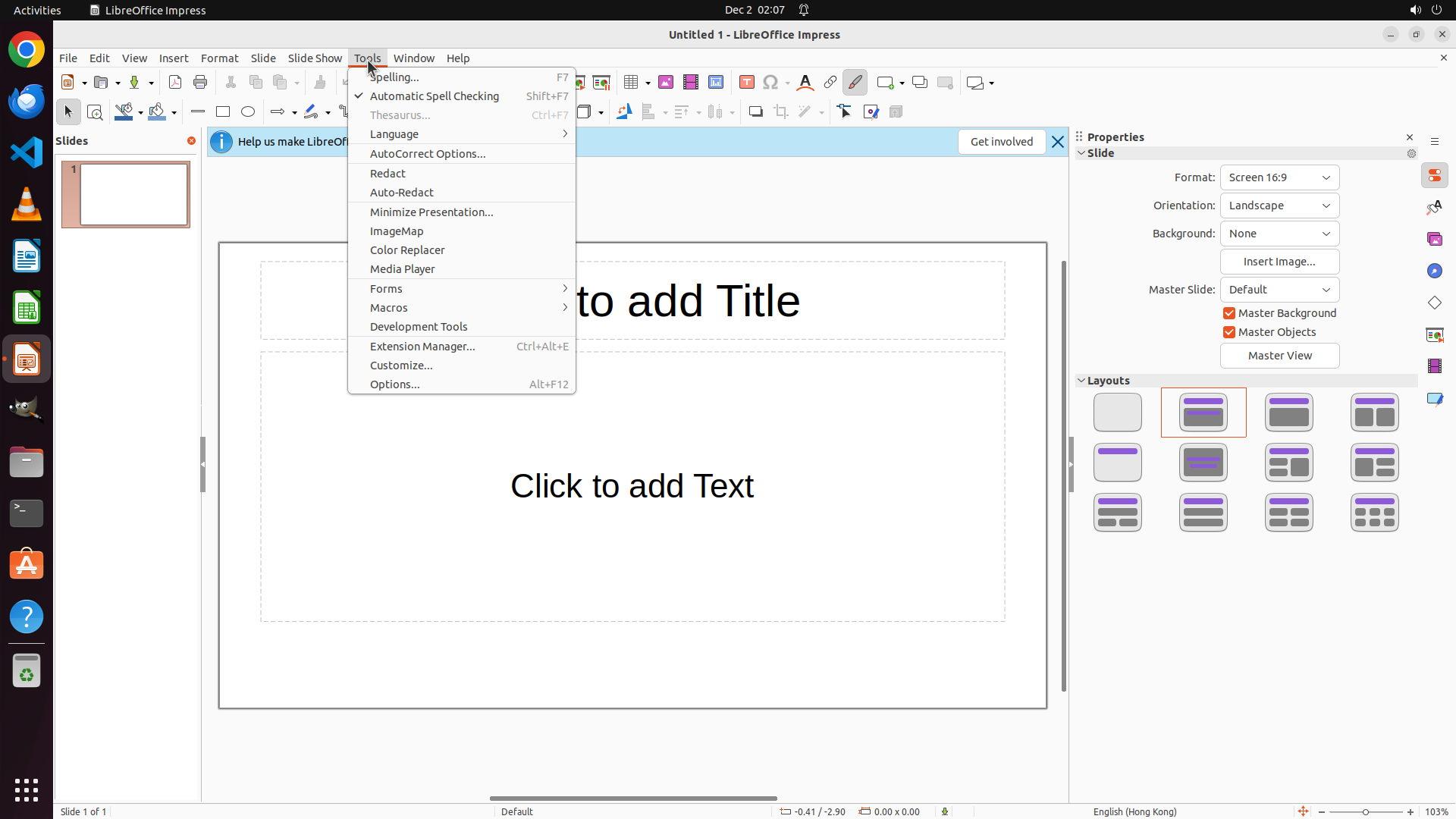}
  \caption{\ours (Step 3)}
\end{subfigure}
\begin{subfigure}{0.32\linewidth}
  \centering
  \includegraphics[width=\linewidth]{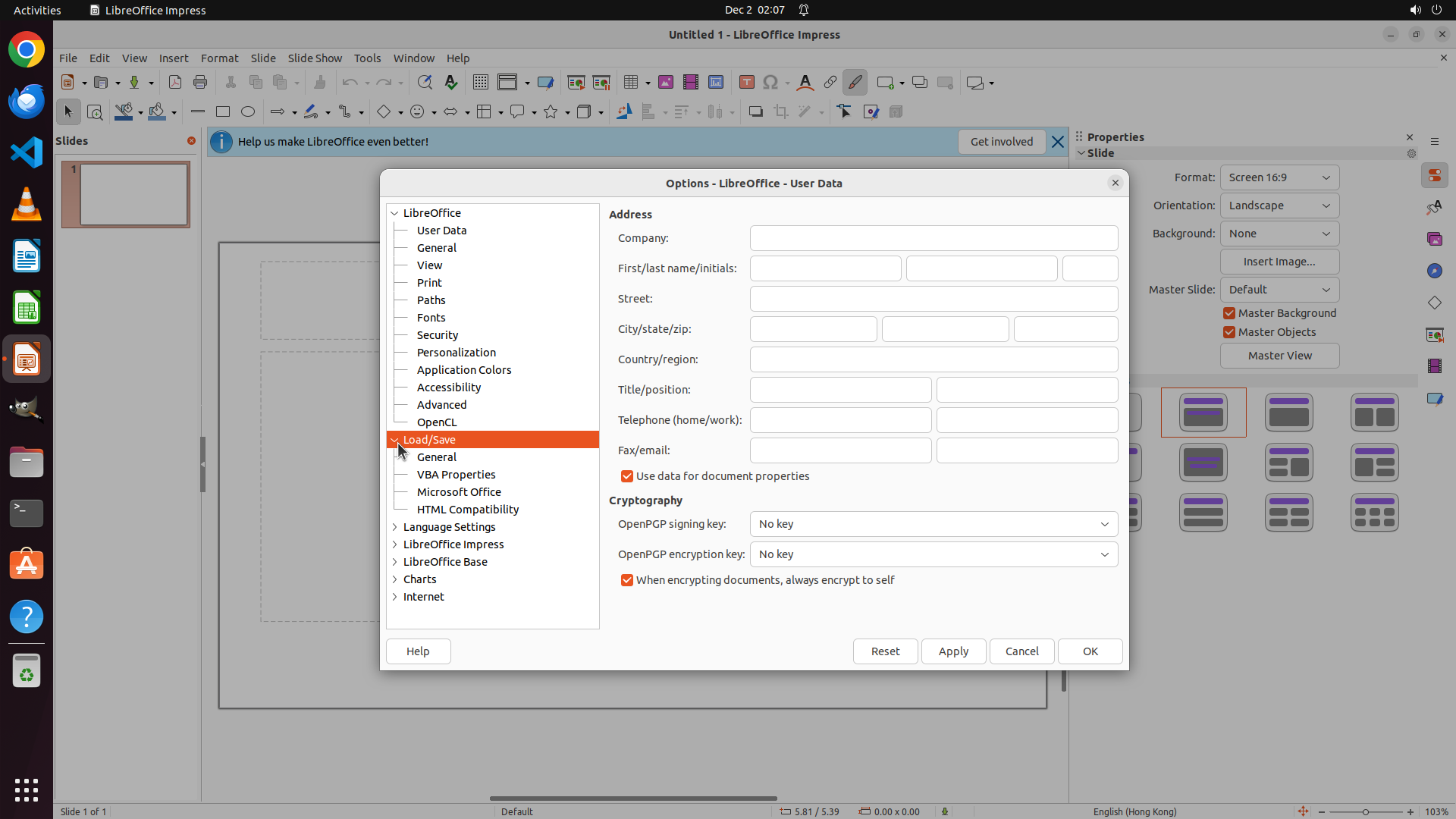}
  \caption{\ours (Step 7)}
\end{subfigure}
\begin{subfigure}{0.32\linewidth}
  \centering
  \includegraphics[width=\linewidth]{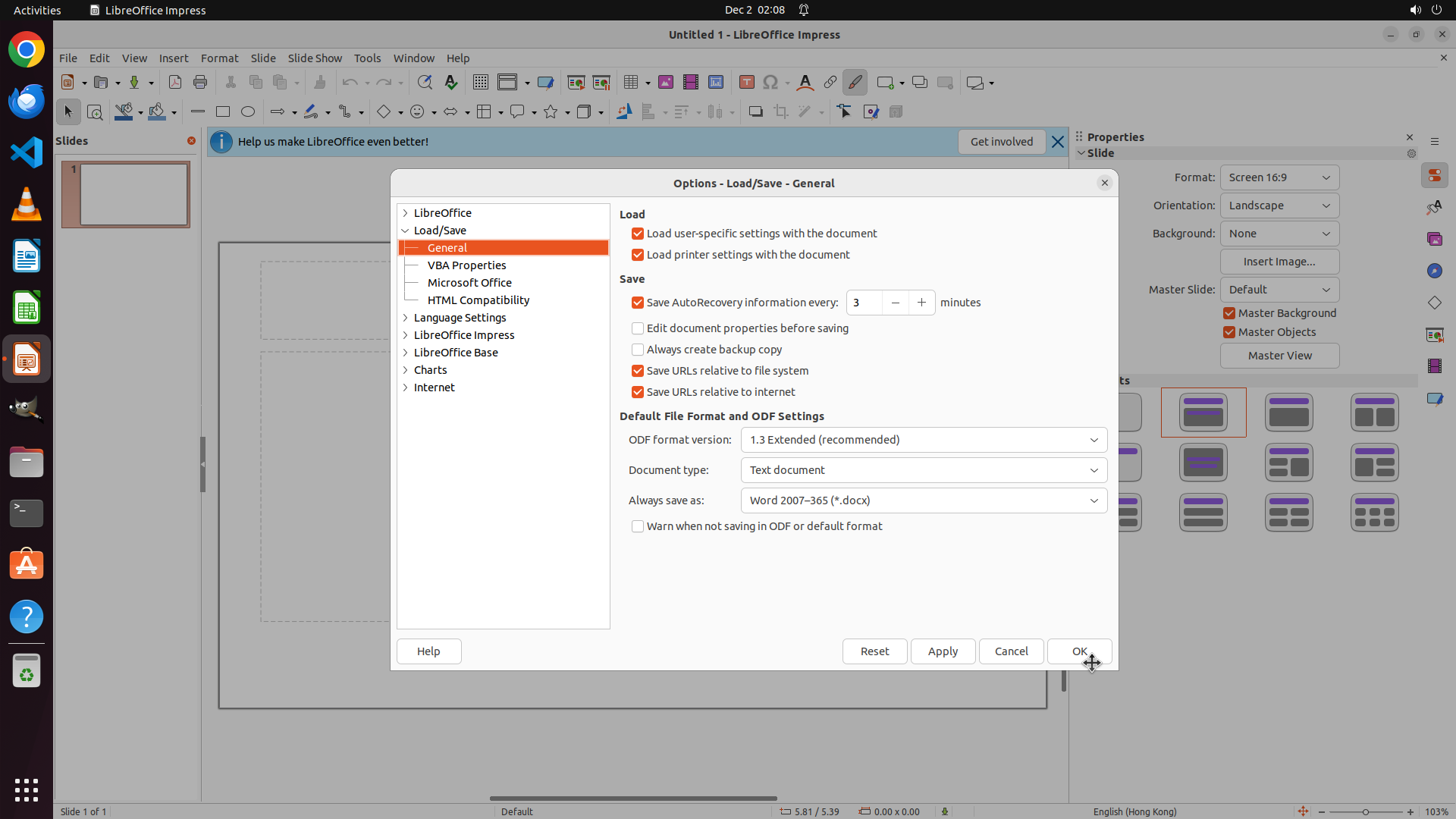}
  \caption{\ours (Step 29)}
\end{subfigure}

\begin{subfigure}{0.32\linewidth}
  \centering
  \includegraphics[width=\linewidth]{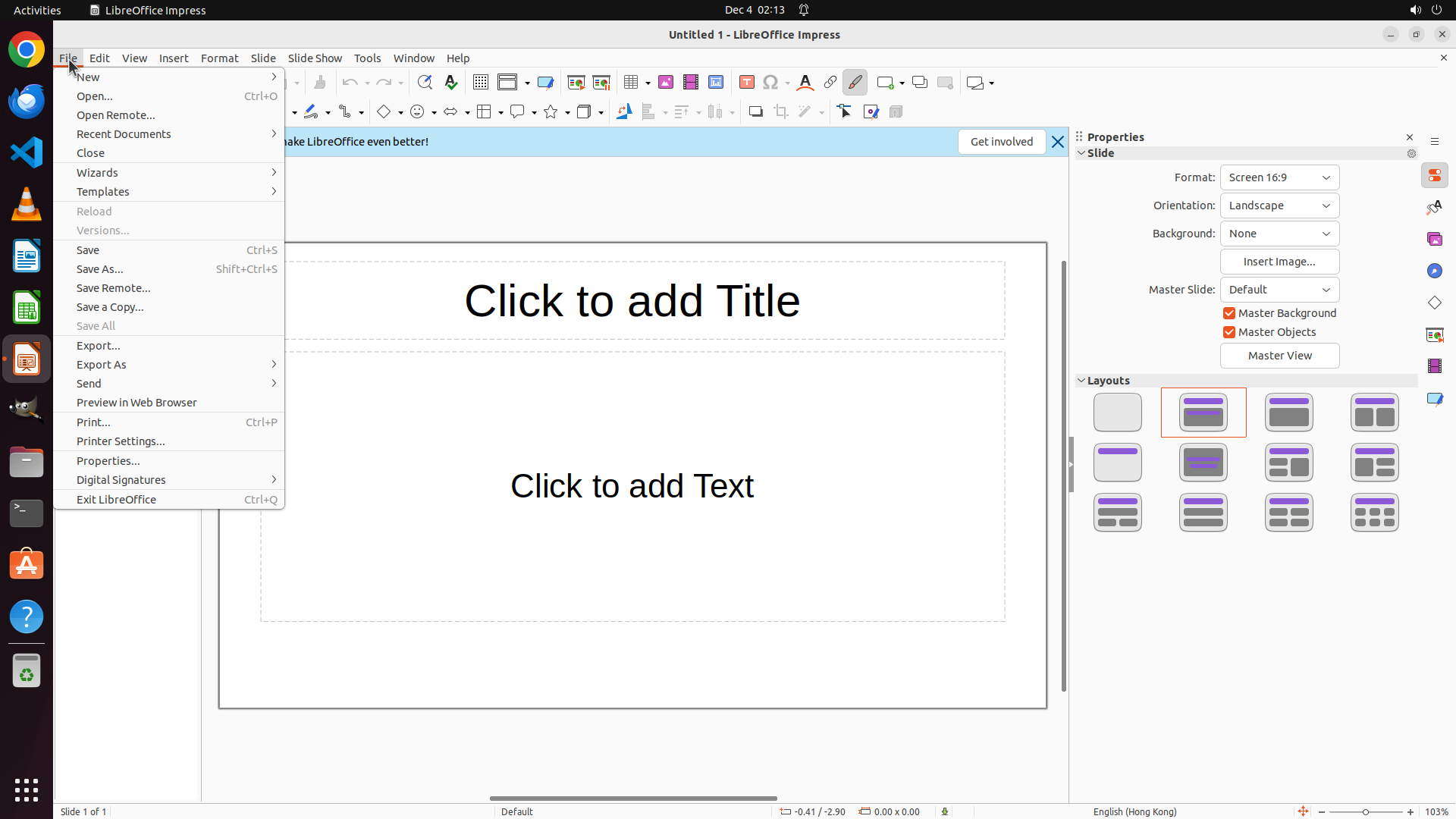}
  \caption{Human-Data (Step 1)}
\end{subfigure}
\begin{subfigure}{0.32\linewidth}
  \centering
  \includegraphics[width=\linewidth]{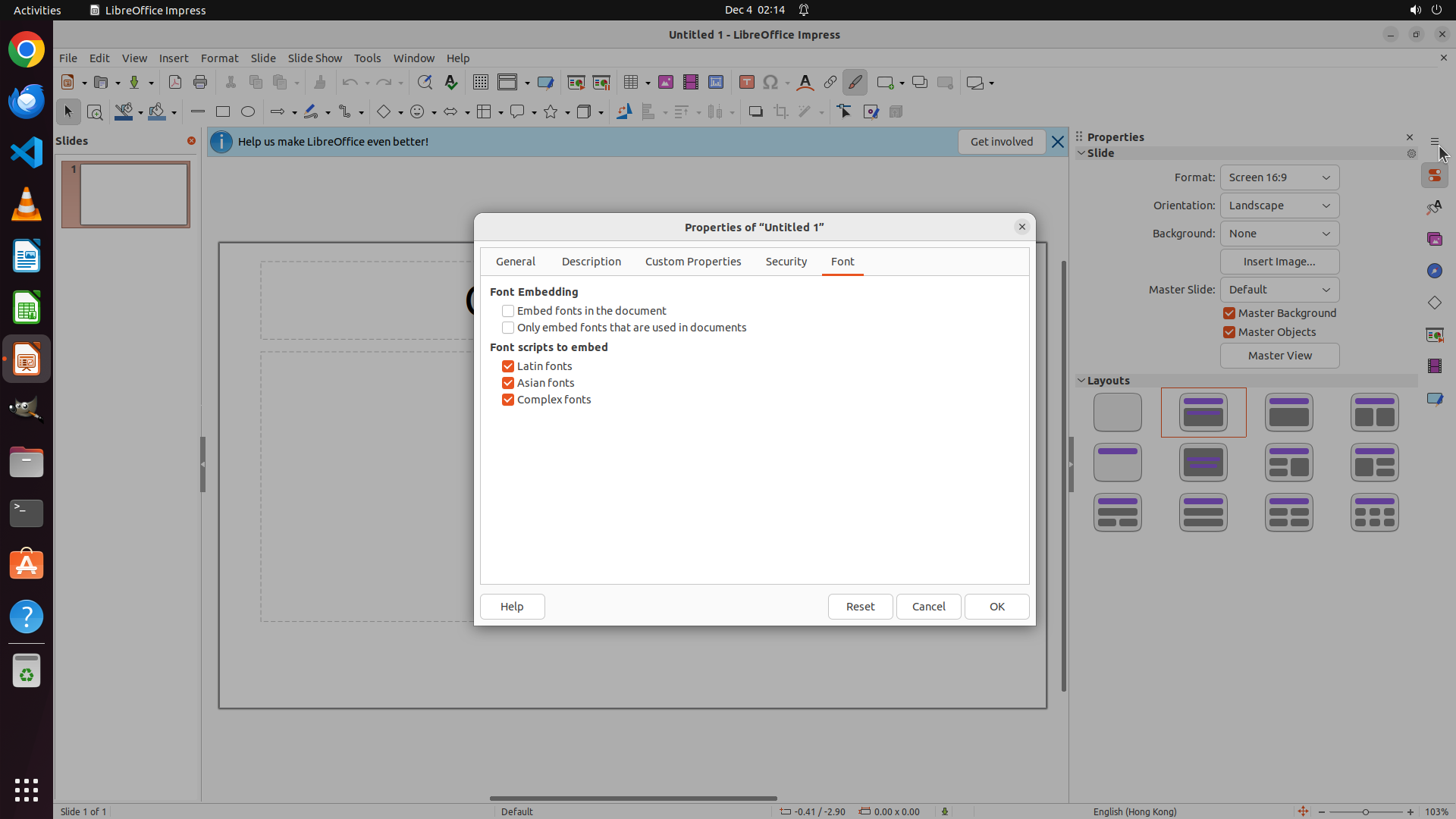}
  \caption{Human-Data (Step 3)}
\end{subfigure}
\begin{subfigure}{0.32\linewidth}
  \centering
  \includegraphics[width=\linewidth]{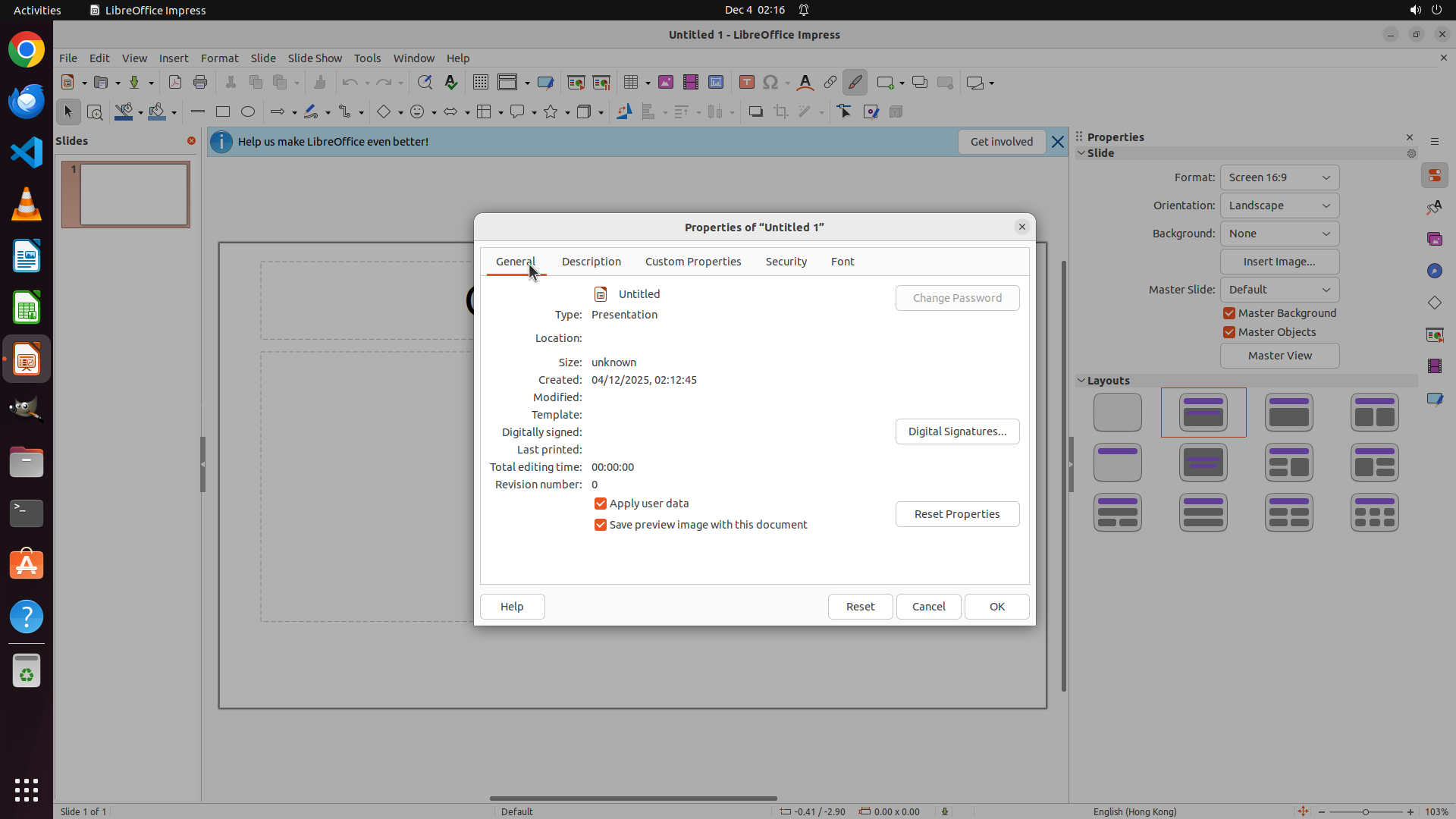}
  \caption{Human-Data (Step 19)}
\end{subfigure}

\begin{subfigure}{0.32\linewidth}
  \centering
  \includegraphics[width=\linewidth]{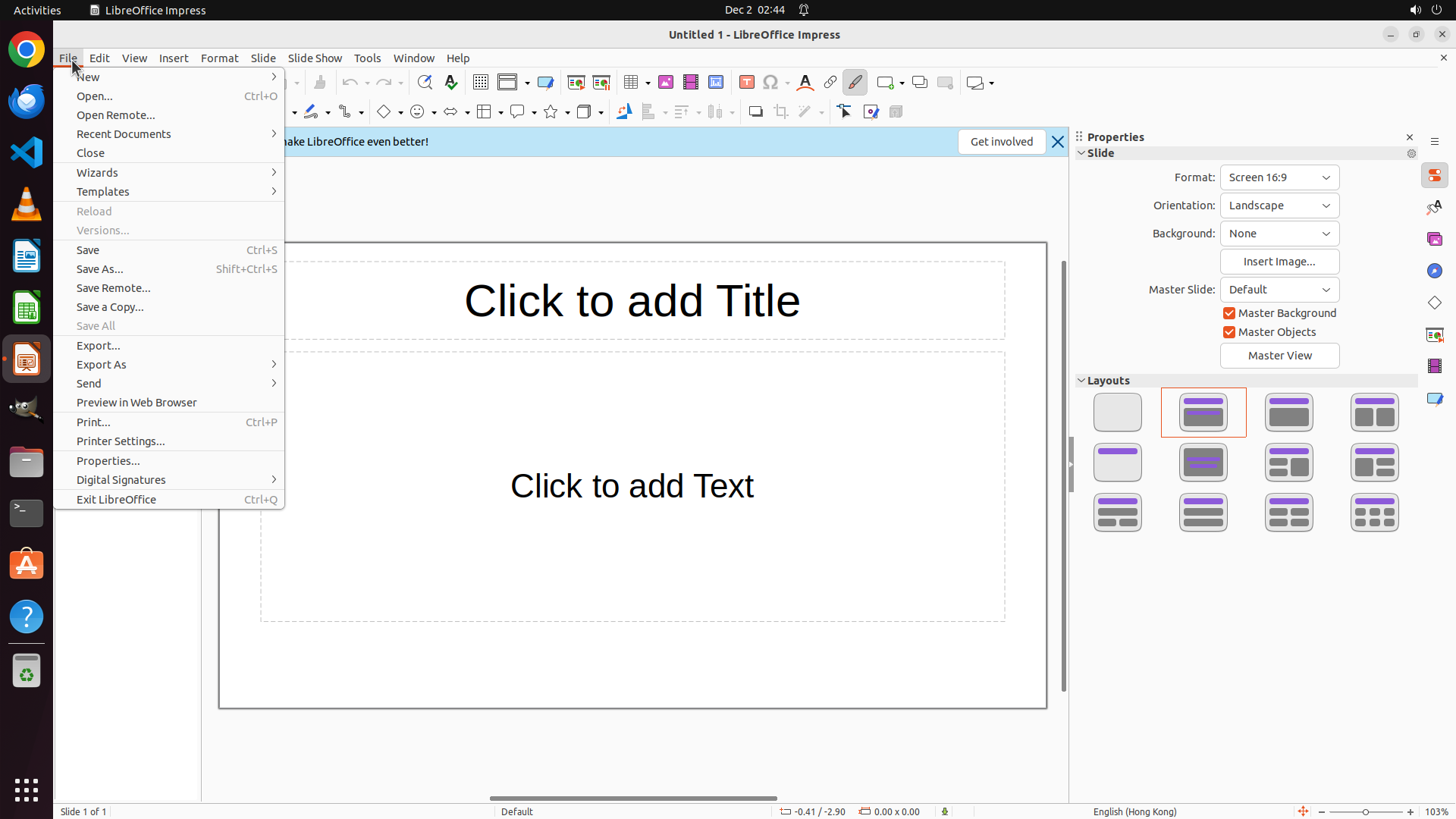}
  \caption{Task-Driven (Step 1)}
\end{subfigure}
\begin{subfigure}{0.32\linewidth}
  \centering
  \includegraphics[width=\linewidth]{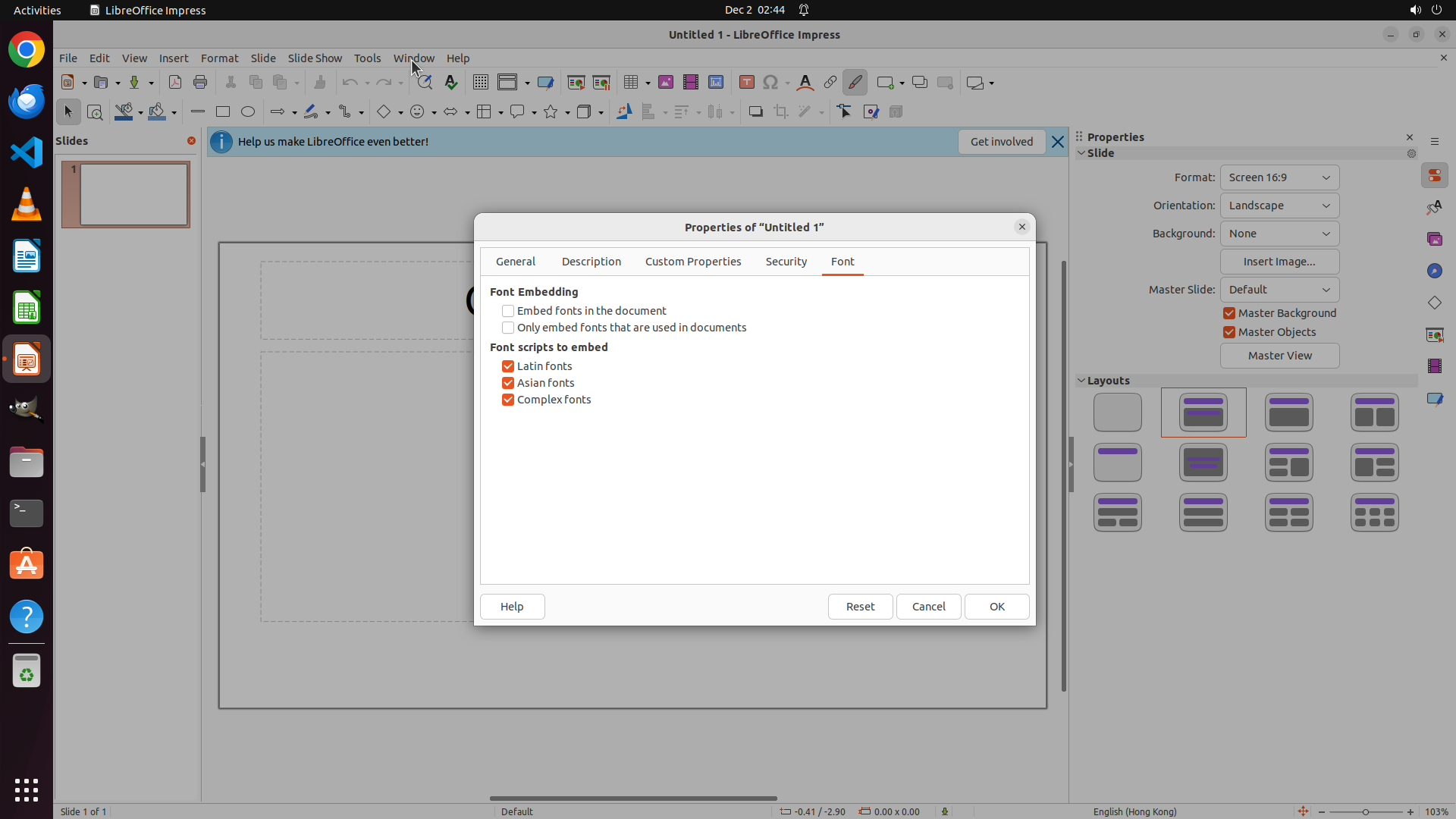}
  \caption{Task-Driven (Step 3)}
\end{subfigure}
\begin{subfigure}{0.32\linewidth}
  \centering
  \includegraphics[width=\linewidth]{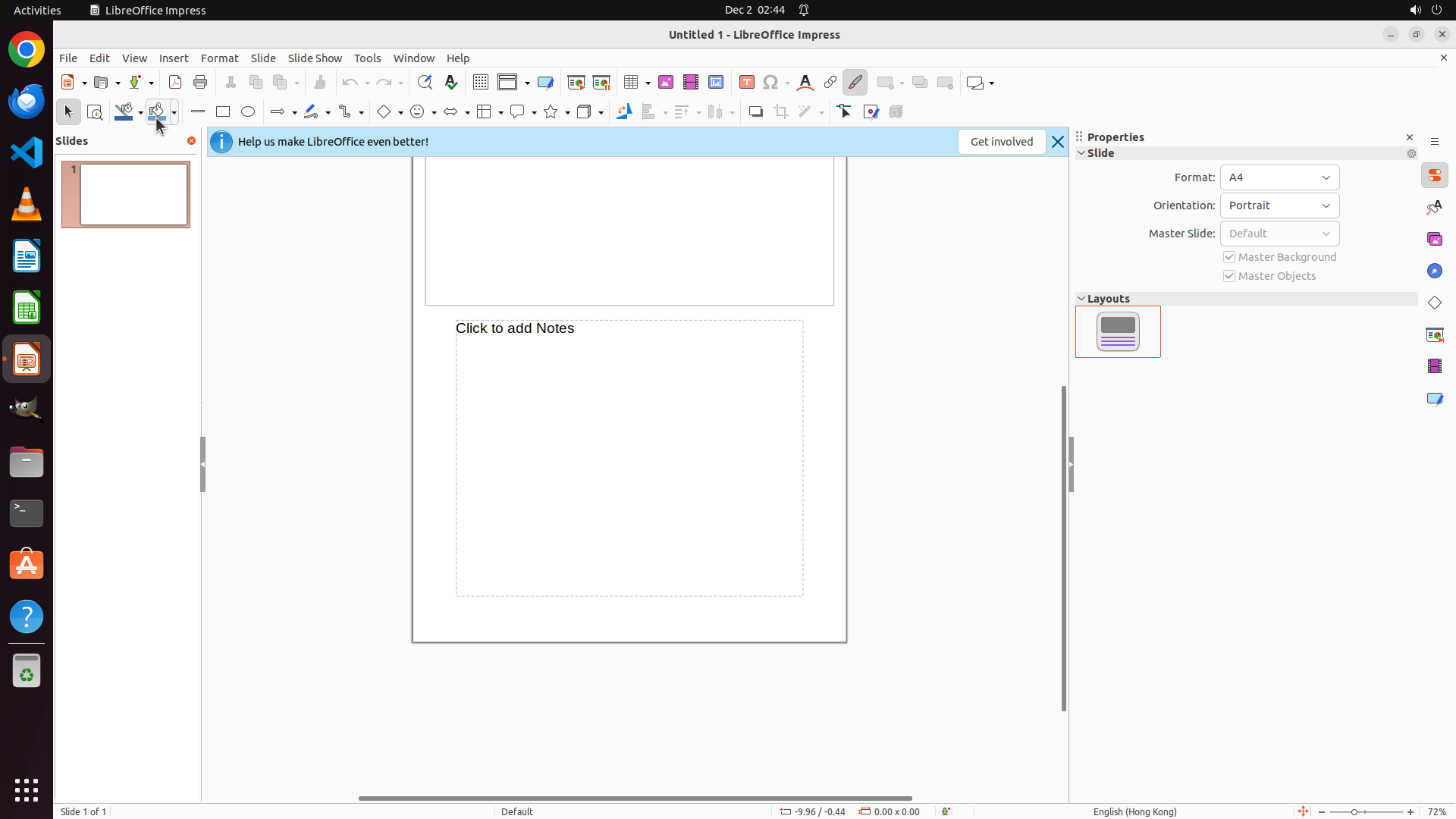}
  \caption{Task-Driven (Step 11)}
\end{subfigure}

\caption{Qualitative comparison on a representative long-horizon settings task (enable LibreOffice auto-save every 3 minutes). \ours correctly navigates the nested \texttt{Tools $\rightarrow$ Options} panels and applies the target setting, while the Human-Data baseline loops within an incorrect submenu and the Task-Driven baseline diverges into a wrong settings page and fails to recover.}

\label{fig:qualitative_comparison}
\end{figure*}

Shown in \autoref{fig:qualitative_comparison}.

\section{Training Details}
\label{app:training_details}

\paragraph{Fair-budget training.}
To ensure a fair comparison, we train all methods with identical trajectory budgets on each benchmark:
$1{,}000$ trajectories per method on OSWorld and $600$ trajectories per method on WindowsAgentArena.
Across methods, trajectories have comparable horizons (about $18$ steps on average).

\paragraph{Training framework.}
We implement training in \textsc{PyTorch} using HuggingFace \texttt{transformers} (\texttt{Trainer}/\texttt{TrainingArguments}) with \texttt{accelerate} for distributed execution, and DeepSpeed ZeRO-3 for memory efficiency (with CPU parameter offloading).
We enable gradient checkpointing and optionally use FlashAttention-2 when supported.

\paragraph{Precision and hardware.}
All models are fully fine-tuned on $4\times$ NVIDIA H200 GPUs.
We train in mixed precision with FP16 by default; models are initialized with \texttt{torch\_dtype=float16}.

\paragraph{Input format and context.}
Each supervised example corresponds to predicting the next step conditioned on the current GUI state and interaction context.
We always include the current screenshot plus \textbf{two preceding screenshots} for all backbones.
We use the original $1920\times1080$ screenshots, and the vision encoder uses patch size 16.

\paragraph{Optimization hyperparameters.}
We use AdamW (\texttt{adamw\_torch}) with $\beta_1=0.9$, $\beta_2=0.95$, $\epsilon=10^{-7}$, weight decay $0.01$, and learning rate $5\times10^{-6}$.
We apply a linear learning-rate schedule with 30 warmup steps and gradient clipping with max grad norm $1.0$.

\paragraph{Batching and training duration.}
We use per-device micro-batch size $1$ and accumulate gradients to reach a global batch size of $16$ across $4$ GPUs.
We train for 1 epoch over the shuffled dataset.

\section{Annotator Information}
\label{app:annotators}

Human annotators were used only for (i) validating candidate seed trajectories (\S\ref{sec:data-generation}) and (ii) auditing 100 automatically verified synthetic trajectories (\S\ref{sec:data_analysis}). Annotators were English-fluent adults recruited from a contractor pool and familiar with common desktop applications. They performed offline verification by inspecting the task instruction together with the recorded trajectory data (screenshots and actions), and then judging whether the final state satisfies the task; for seed validation they also checked for obvious redundant detours and harmful side effects (e.g., deleting unrelated files or changing global settings). For the audit, annotators labeled success/failure without access to the verifier prediction. No personal data or real user accounts were accessed.

\end{document}